\documentclass[sigconf,screen]{acmart}
\usepackage{wrapfig}
\usepackage{subfig}
\usepackage[ruled]{algorithm2e}

\SetKwInput{KwInput}{Input}                % Set the Input
\SetKwInput{KwOutput}{Output}              % set the Output
\SetKwInput{KwTo}{to} 
\AtBeginDocument{%
  \providecommand\BibTeX{{%
    \normalfont B\kern-0.5em{\scshape i\kern-0.25em b}\kern-0.8em\TeX}}}

\newcommand{\squishlist}{
 \begin{list}{$\bullet$}
  { \setlength{\itemsep}{0pt}
     \setlength{\parsep}{3pt}
     \setlength{\topsep}{3pt}
     \setlength{\partopsep}{0pt}
     \setlength{\leftmargin}{1.5em}
     \setlength{\labelwidth}{1em}
     \setlength{\labelsep}{0.5em} } }

\newcommand{\squishlisttwo}{
 \begin{list}{$\bullet$}
  { \setlength{\itemsep}{0pt}
     \setlength{\parsep}{0pt}
    \setlength{\topsep}{0pt}
    \setlength{\partopsep}{0pt}
    \setlength{\leftmargin}{2em}
    \setlength{\labelwidth}{1.5em}
    \setlength{\labelsep}{0.5em} } }

\newcommand{\squishend}{
  \end{list}  }

\renewcommand\footnotetextcopyrightpermission[1]{} % removes footnote with conference information in first column
\begin{document}

%\title{MUSSEL: A MUlti-task Self-SupervisEd Learning Framework for Recommenders}
\title{Self-supervised Learning for Large-scale Item Recommendations}

\author{
Tiansheng Yao, Xinyang Yi, Derek Zhiyuan Cheng, Felix Yu, Ting Chen, Aditya Menon}
% \email{{tyao, xinyang, zcheng, felixyu, iamtingchen, adityakmenon}@google.com}

\author{
Lichan Hong, Ed H. Chi, Steve Tjoa, Jieqi (Jay) Kang, Evan Ettinger}
% \email{{lichan, edchi, stevetjoa, jaykang, eettinger}@google.com}
\affiliation{%
Google Inc.
\country{United States}}

\begin{abstract}
Large scale recommender models find most relevant items from huge catalogs, and they play a critical role in modern search and recommendation systems. To model the input space with large-vocab categorical features, a typical recommender model learns a joint embedding space through neural networks for both queries and items from user feedback data. However, with millions to billions of items in the corpus, users tend to provide feedback for a very small set of them, causing a power-law distribution. This makes the feedback data for long-tail items extremely sparse.

Inspired by the recent success in self-supervised representation learning research in both computer vision and natural language understanding, we propose a multi-task self-supervised learning (SSL) framework for large-scale item recommendations. The framework is designed to tackle the label sparsity problem by learning better latent relationship of item features. Specifically, SSL improves item representation learning as well as serving as additional regularization to improve generalization. Furthermore, we propose a novel data augmentation method that utilizes feature correlations within the proposed framework.

We evaluate our framework using two real-world datasets with 500M and 1B training examples respectively. Our results demonstrate the effectiveness of SSL regularization and show its superior performance over the state-of-the-art regularization techniques. We also have already launched the proposed techniques to a web-scale commercial app-to-app recommendation system, with significant 
improvements top-tier business metrics demonstrated in A/B experiments on live traffic. Our online results also verify our hypothesis that our framework indeed improves model performance even more on slices that lack supervision.
\end{abstract}

%\keywords{Recommendation, Deep Learning, Self-supervised Learning}

\maketitle
\pagestyle{plain}

\section{Introduction} \label{sec:intro}
Recently, neural-net models have emerged to the main stage of modern recommendation systems throughout the industry (see, e.g., \cite{He2017, maxim2019, zhe19watchnext, yi2019samplingbias}), and academia (\cite{chris_kdd_2018, npr_wsdm_2018}). Compared to conventional approaches like matrix factorization \cite{matrix_factorization, advanced_cf, focused_learning_2017}, gradient boosted decision trees \cite{xgboost, mounia_www_2019}, and logistic regression based recommenders \cite{fb_adkdd_14}, these deep models handle categorical features more effectively. They also enable more complex data representations, and introduce more non-linearity to better fit the complex data for recommenders.

A particular recommendation task we focus on in this paper is to identify the most relevant items given a query from a huge item catalog. This general problem of \textsl{large-scale item recommendations} has been widely adopted in various applications. Depending on the type of the query, a recommendation task could be: (i) personalized recommendation: when the query is a user; (ii) item to item recommendation: when the query is also an item; and (iii) search: when the query is a piece of free text. To model the interactions between a query and an item, a well-known approach leverages \textsl{embedding-based neural networks}. The recommendation task is typically formulated as an extreme classification problem \cite{paul2016} where each item is represented as a dense vector in the output space.

\begin{figure}
    \centering
    \includegraphics[width=0.8\columnwidth]{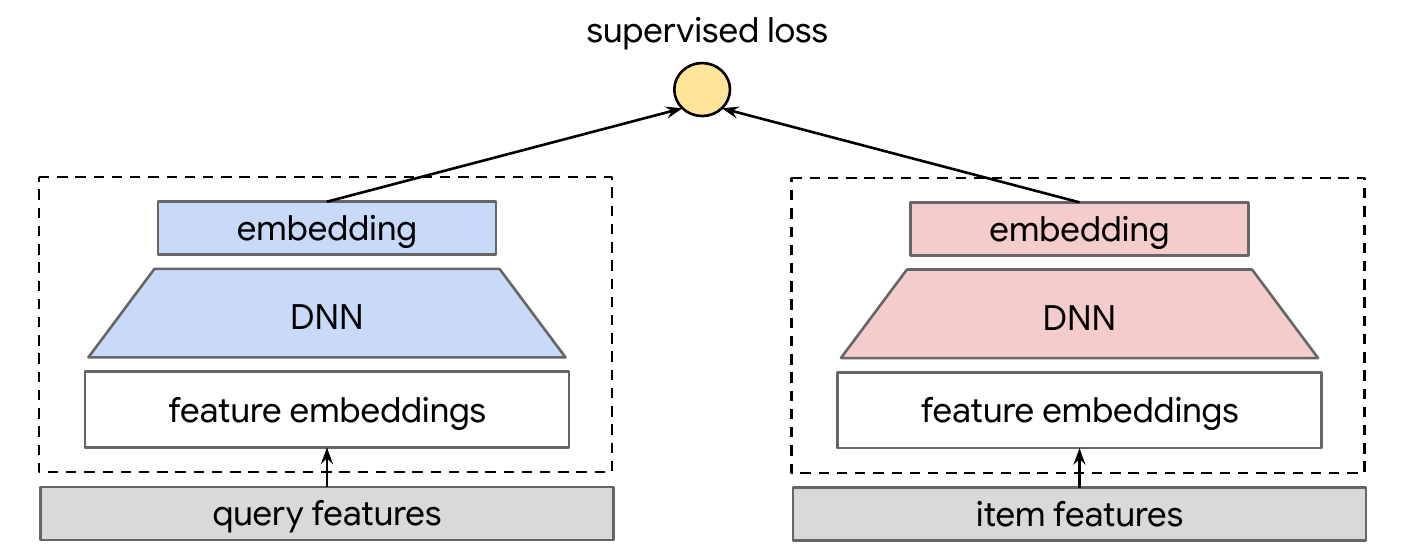}
    \caption{Model architecture: Two-tower DNN with query and item representations.}
    \vspace{-10pt}
    \label{fig:two_tower_dnn}
\end{figure}

This paper is focused on the \textsl{two-tower DNNs} (see Figure \ref{fig:two_tower_dnn}), popular in many real-world recommenders (see e.g. \cite{krichene2018efficient, yi2019samplingbias}). In this architecture, a neural network encodes a set of item features into an embedding thus making it applicable even for indexing cold-start items. Moreover, the two-tower DNN architecture enables efficient serving for a large corpus of items in real time, by converting the top-k nearest neighbor search problem to Maximum-Inner-Product-Search (MIPS) \cite{Cohen1997} that is solvable in sublinear complexity.

Embedding-based deep models typically have large amount of parameters because they are built with high-dimensional embeddings that represent high cardinality sparse features such as topics or item IDs. 
In many existing literature, the loss functions for training these models are formulated as a supervised learning problem. The supervision comes from the collected labels (e.g., clicks). Modern recommendation systems collect billions to trillions of footprints from users, providing huge amount of training data for building deep models. However, when it comes to modeling a huge catalogue of items in the order of millions (e.g., songs and apps \cite{Mark2010}) to even billions (e.g., videos on YouTube \cite{paul2016}), data could still be highly sparse for certain slices due to:

\squishlist
\item \textbf{Highly-skewed data distribution}: The interaction between queries and items are often highly skewed in a power-law distribution \cite{Milojevic2010}. So a small set of the popular items gets most of the interactions. This will always leave the training data for long-tail items very sparse.
\item \textbf{Lack of explicit user feedback}: Users often provide lots of positive feedback implicitly like clicks and thumb-ups. However, they are much less likely to provide explicit feedback like item ratings, feedback for user happiness, and relevance scores.
\squishend

%\begin{itemize}
%    \item \textbf{Highly-skewed data distribution}: The interaction between queries and items are often highly skewed in a power-law distribution \cite{Milojevic2010}, meaning only a small percentage of the items get most of the interactions.
%    \item \textbf{Gigantic corpus}: The catalogue of items is often in the order of millions (e.g., songs, apps) \cite{Mark2010} to billions (e.g., posts in Facebook, videos on YouTube) \cite{paul2016}.
%    \item \textbf{Lack of explicit user feedback}: Although modern recommendation systems in the industry collect hundreds of billions of implicit user actions like clicks and thumb-ups, these systems still lack high quality user explicit feedback like item ratings, feedback for user happiness, and relevance scores.
%\end{itemize}

Self-supervised learning (SSL) offers a different angle to improve deep representation learning via unlabeled data. The basic idea is to enhance training data with various data augmentations, and supervised tasks to predict or reconstitute the original examples as auxiliary tasks. Self-supervised learning has been widely used in the areas of Compute Vision (CV) \cite{larsson2016learning, noroozi2016unsupervised, gidaris2018unsupervised} and Natural Language Understanding (NLU) \cite{devlin2018bert, Lan2020ALBERT}. An example work \cite{larsson2016learning} in CV proposed to rotate images at random, and train a model to predict how each augmented input image was rotated. In NLU, masked language task was introduced in the BERT model, to help improve pre-training of language models. Similarly, other pre-training tasks like predicting surrounding sentences and linked sentences in Wikipedia articles have also been used in improving dual-encoder type models in NLU \cite{pre-training}.
Compared to conventional supervised learning, self-supervised learning provides complementary objectives eliminating the pre-requisite of collecting labels manually. In addition, SSL enables autonomous discovery of good semantic representations by exploiting the internal relationship of input features.

%\felix{Shall we add a sentence or two on the benefits? in comparison with conventional supervised learning, self-supervised learning provides..... }

Despite the wide adoption in computer vision and natural language understanding, the application of self-supervised learning in the field of recommendation systems is less well studied. The closest line of research studies a set of regularization techniques \citep{xu2017spreadout, krichene2018efficient, guo2019glass}, which are designed to force learned representations (i.e., output layer (embeddings) of a multi-layer perception), of different examples to be farther away from each other, and spread out in the entire latent embedding space. Although sharing similar spirit with SSL, these techniques do not explicitly construct SSL tasks. In contrast to models in CV or NLU applications, recommendation model takes extremely \text{sparse} input where high cardinality categorical features are one-hot (or multi-hot) encoded, such as the item IDs or item categories \cite{maxim2019}. These features are typically represented as learnable embedding vectors in deep models. As most models in computer vision and NLU deal with dense input, existing methods for creating SSL tasks are not directly applicable to the sparse models in recommendation systems. More recently, a line of research studies self-supervised learning improving sequential user modeling in recommendations \cite{ma2020kdd, Xin2020SelfSupervisedRL, Zhou2020S3RecSL}. Different from these works, this paper focuses on item representation learning, and shows how SSL can help improve generalization in the context of long-tail item distribution. Moreover, in contrast to using SSL on a certain sequential user feature, we design new SSL tasks and demonstrate their effectiveness for learning with a set of heterogeneous categorical features, which we believe is a more general setup for other types of recommendation models such as multitask ranking models (e.g., \cite{zhe19watchnext}).

In this paper, we propose to leverage self-supervised learning based auxiliary tasks to improve item representations, especially with long-tail distributions and sparse data. Different from CV or NLU applications, input space of recommendation model is highly \textsl{sparse} and represented by a set of categorical features (e.g. item ids) with large cardinality. For such sparse models, we propose a new SSL framework, where the key idea is to: (i) augment data by masking input information; (ii) encode each pair of augmented examples by a two-tower DNN; and (iii) apply a contrastive loss to learn representations of augmented data. The goal of contrastive learning is to let augmented data from the same example be discriminated against others. Note that the two-tower DNN for contrastive learning and the one for encoding query and item can share a certain amount of model parameters. See more details in Section \ref{sec:method}.

Our contribution is four-fold:
% TYAO: I remove this part since it's redundant
% We propose to feed augmented example features to ach tower, and use a contrastive prediction task to identify the pair of augmented examples from the same original before data augmentation.

% ZCHENG: these were highly redundant compared to the bullet points below.
% We offer two simple methods for example augmentation -- feature masking and feature dropout -- which are easy to use for sparse models for recommenders. To leverage the proposed SSL task for main prediction task relying on supervised labels, we also offer a multi-task learning approach where these two tasks are jointly optimized.

\squishlist
\item \textbf{SSL framework}: We present a model architecture agnostic self-supervised learning framework for sparse neural models in recommendations. The auxiliary self-supervised loss and the primary supervised loss are jointly optimized via a multi-task learning framework. We focus on using this framework for efficiently scoring a large corpus of items, which is also known as item retrieval in two-stage recommenders \cite{paul2016}. We believe it would also shed light on designing SSL for other types of models such as ranking models \cite{heng16}.

\item \textbf{Data augmentation}: We propose a novel data augmentation method that exploits feature correlations and are tailored for heterogeneous categorical features that are common in recommender models.

\item \textbf{Offline experiments}: On one public dataset and one industry scale dataset for recommendation systems, we demonstrate that introducing SSL as an auxiliary task can significantly improve model performance, especially when labels are scarce. Comparing to the state-of-art non-SSL regularization techniques \citep{xu2017spreadout, krichene2018efficient, guo2019glass}, we demonstrate that SSL consistently performs better, and improves model performance even when non-SSL regularization does not bring any additional gains.

\item \textbf{Live experiment in a web-scale recommender}: We have launched the proposed SSL technique in a fairly strong two-tower app-to-app recommendation model in a large-scale real-world system. Live A/B testing shows significantly improved top-tier metrics. We especially see bigger improvements for slices without much supervision.

\squishend
\vspace{-8pt}

\section{Related Work}

\label{sec:related_work}
\paragraph{Self-supervised Learning and Pre-training}
Various unsupervised and self-supervised learning tasks have been studied in the computer vision community. The closest line of research is SimCLR~\cite{chen2020simple} which also utilizes self-supervised learning and contrastive learning for visual representation learning. Different with SimCLR and other works~\cite{chen2020big, s4l} in vision, here we propose augmentations that are more suitable and tailed for categorical features for recommendations, instead of relying on image-specific augmentations such as image cropping, rotation and color distortion. In addition, the proposed framework does not require multi-stage training schedules (such as pre-training then fine-tuning).
\cite{kolesnikov2019revisiting}.

%Popular SSL tasks include: (i) predicting image rotations \cite{gidaris2018unsupervised}; (ii) predicting relative patch locations \cite{noroozi2016unsupervised}; (iii) predicting next video frames \cite{srivastava2015unsupervised} ; and (iv) leveraging contrastive loss \cite{dosovitskiy2014discriminative} etc. In natural language understanding, SSL like pre-training tasks such as next sentence/word prediction and masked-LM have been widely used \cite{devlin2018bert}. 

In NLU, for dual-encoder models, \cite{pre-training} shows that pre-training tasks better aligned with the final task are more helpful than generic tasks such as next sentence prediction and masked-LM. The pre-training tasks are designed to leverage large-scale public NLU content, such as Wikipedia. In this paper, we also use the dual-encoder model architecture. Different from the above, the proposed self-supervision tasks do not require the use of a separate data source. 

%In computer vision, the closest line of research is SimCLR~\cite{chen2020simple} which also utilizes self-supervised learning and contrastive learning for visual representation learning. Different with SimCLR and other works~\cite{chen2020big, s4l} in vision, here we propose augmentations that are more suitable and tailed for categorical features for recommendations, instead of relying on image-specific augmentations such as image cropping, rotation and color distortion. In addition, the proposed framework does not require multi-stage training schedules (such as pre-training then fine-tuning).

\noindent \paragraph{Spread-out Regularization}
\citet{xu2017spreadout} and \citet{wu2018unsupervised} use spread-out regularization for improving generalization of deep models. Specifically, in \cite{xu2017spreadout}, a regularization promoting separation between random instances is shown to improve training stability and generalization. In \cite{wu2018unsupervised}, one objective is to train a classifier treating each instance as its own class, therefore promoting large instance separations in the embedding space. Both the above approaches are studied for computer vision applications.

% Google Drawing location:
% https://docs.google.com/drawings/d/13WIT6gO4fPzwmvNN8HvM8Sg0kDdQzb-fEEzX5qjnrqk/edit

\noindent \paragraph{Neural Recommenders} Deep learning has led to many successes in building industry-scale recommender systems such as video suggestion \cite{paul2016}, news recommendation \cite{Okura2017EmbeddingbasedNR}, and visual discovery in social networks \cite{Liu2017, Zhai2017}. For large-scale item retrieval, two-tower models with separate query and item towers are widely used due to its high efficiency for serving. The recommendations are typically computed by a dot-product between query and item embeddings so that finding top-k items can be converted to MIPS problem \cite{Cohen1997} with sublinear time complexity. One popular factorized structure is softmax-based multi-class classification model. The work in \cite{paul2016} treats the retrieval task as an extreme multi-class classification trained with multi-layer perceptron (MLP) model using sampled softmax as loss function. Such models leverages item ID as the only item feature, suffering from cold-start issue. More recently, a line of research \cite{krichene2018efficient, yi2019samplingbias} considers applying two-tower DNNs on retrieval problems, which is also known as dual-encoder \cite{gillick2018endtoend, yang-etal-2018-learning}, where item embeddings are constructed by a MLP from ID and other categorical metadata features. The self-supervised approach proposed is applicable to both ranking and retrieval models. In this paper we focus on using SSL for retrieval models, particularly, on improving item representations in two-tower DNNs.

%To handle the challenge of scoring a large number of items, many recommender systems follow a two-stage design that first relies on a \textit{candidate generation} (retrieval) model to find thousands of relevant items given a query, and then uses a finer-granularity \textit{ranking model}  to identify the best items to show to the user \cite{heng16, zhe19watchnext}. 

\noindent \paragraph{Self-supervised Learning in Sequential Recommendations} In recommender systems, a line of research has been recently studied for utilizing self-supervised learning for sequential recommendation. Self-supervised learning tasks are designed to capture information among user history~\cite{Zhou2020S3RecSL} and learn more robust disentangled user representation~\cite{ma2020kdd} in user sequential recommendation. Moreover, \citeauthor{Xin2020SelfSupervisedRL} shows combining SSL with reinforcement learning is effective to capture long-term user interest in sequential recommendation. Different from the above, our proposed SSL framework is focusing on improving item representation with long-tail distributions. The proposed SSL tasks do not require modeling sequential information and are generally applicable to deep models with heterogeneous categorical features.

\section{Method}
%\xinyang{nit: the title looks similar to the first subtitle below.}
%updated the subtitle to be just Framework.
\label{sec:method}
We present our framework of self-supervised learning for deep neural net models for recommenders using large-vocab categorical features. Particularly, a general self-supervised learning framework is introduced in Section \ref{method:ssl_framework}. In Section \ref{method:ssl_tasks}, we present a data augmentation method to construct SSL tasks and elaborates on their connections to spread-out regularization. Finally, in Section \ref{method:ssl_multitask}, we describe how to use SSL to improve factorized models (i.e., two-tower DNNs as shown in Figure \ref{fig:two_tower_dnn}), via a multi-task learning framework.

%It is worth noting that the SSL framework we propose is model architecture agnostic. In this paper, we focus on two-tower models, however the same learning should apply to other model architectures like matrix factorization and multilayer perceptron.

\begin{figure}
    \centering
    \includegraphics[width=1\columnwidth]{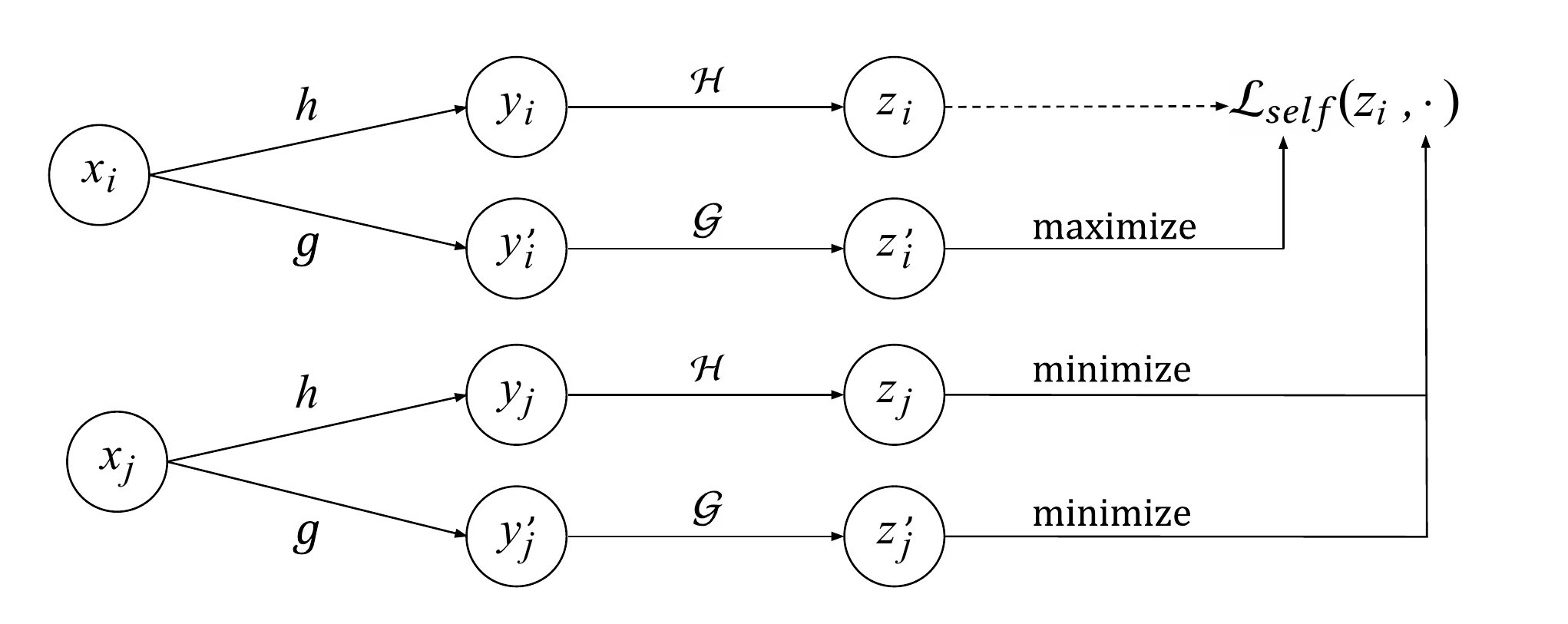}
    \caption{Self-supervised learning framework illustration. Two data augmentations $h$ and $g$ are applied to the input; Encoders $\mathcal{H}$ and $\mathcal{G}$ are applied to the augmented examples $\mathbf{y}_i$ and $\mathbf{y}_i'$ to generate embeddings $\mathbf{z}_i$ and $\mathbf{z}_i'$. The SSL loss $\mathcal{L}_{self}$ w.r.t. $\mathbf{z}_i$ is optimized towards maximizing the similarity with $\mathbf{z}_i'$ while minimizing the similarity between $\mathbf{z}_j$ and $\mathbf{z}_j'$.}\label{fig:ssl_framework}
    \vspace{-15pt}
\end{figure}

\subsection{Framework} \label{method:ssl_framework}
Inspired by the SimCLR framework \cite{chen2020simple} for visual representation learning, we adopt similar contrastive learning algorithms for learning  representations of categorical features. The basic idea is two folds: first, we apply different data augmentation for the same training example to learn representations; and then use contrastive loss function to encourage the representations learned for the same training example to be similar. Contrastive loss was also applied in training two-tower DNNs (see e.g., \cite{krichene2018efficient, yi2019samplingbias}), although the goal there was to make positive item agree with its corresponding queries.

%\felix{I think there is a distinction on ``example'' and ``item''. Should we clarify EARLY HERE that in this paper, we use items as examples but the proposed method can also be used on queries?} \xinyang{Good point. Since we have similar statement below, here I just clarify example is used to refer item, query or both.}.

We consider a batch of $N$ item examples $x_1,...,x_N$, where $x_i \in \mathcal{X}$ represents a set of features for example $i$. In the context of recommenders, an example indicates a query, an item or a query-item pair. Suppose there are a pair of transform function $h, g: \mathcal{X} \rightarrow \mathcal{X}$ that augment $x_i$ to be $y_i, y_i'$ respectively,
\begin{equation}
y_i \leftarrow h(x_i),~ y_i' \leftarrow g(x_i).
\end{equation}
Given the same input of example $i$, we want to learn different representations $y_i, y_i'$ after augmentation to make sure the model still recognizes that both $y_i$ and $y_i$ represent the same input $i$. In other words, the contrastive loss learns to minimize the difference between $y_i, y_i'$. In the mean time, for different example $i$ and $j$, the contrastive loss maximizes the difference between the representations learned $y_i, y_j'$ after data different augmentations. Let $\mathbf{z}_i, \mathbf{z}_i'$ denote the embeddings of $y_i, y_i'$ after encoded by two neural networks $\mathcal{H}, \mathcal{G}: \mathcal{X} \rightarrow \mathbb{R}^d$, that is

%\felix{I am confused here. I think we should use the SAME neural network.} \xinyang{For feature prediction, they can't be the same as input layer sizes are different.} \felix{but for feature dropout they have to share other layers?} \xinyang{That's true. I discussed the parameter sharing in the paragraph below. Maybe I should explicitly mention they are shared when introducing feature dropout?} \xinyang{added one sentence to each of the tasks in section 3.2 to clarify the parameter sharing of H, G}

\begin{equation}
\mathbf{z}_i \leftarrow \mathcal{H}(y_i),~ \mathbf{z}_i' \leftarrow \mathcal{G}(y_i').
\end{equation}

We treat $(\mathbf{z}_i, \mathbf{z}_i')$ as positive pairs, and $(\mathbf{z}_i, \mathbf{z}_j')$ as negative pairs for $i \ne j$. Let $s(\mathbf{z}_i, \mathbf{z}_j') = \langle \mathbf{z}_i, \mathbf{z}_j' \rangle / (\|\mathbf{z}_i\|\cdot \|\mathbf{z}_j'\|)$. To encourage the above properties, we define the SSL loss for a batch of $N$ examples $\{x_i\}$ as:
\begin{equation} \label{ssl_loss}
\mathcal{L}_{self}(\{x_i\} ; \mathcal{H}, \mathcal{G}) := -\frac{1}{N} \sum_{i \in [N]}\log \frac{\exp{(s(\mathbf{z}_i, \mathbf{z}_i')/
\tau)}}{\sum_{j \in [N]} \exp{(s(\mathbf{z}_i,\mathbf{z}_j')/\tau)}}.
\end{equation}
where $\tau$ is a tunable hyper-parameter for the softmax temperature.
The above loss function learns a robust embedding space such that similar items are close to each other after data augmentation, and random examples are pushed farther away. The overall framework is illustrated in Figure \ref{fig:ssl_framework}.

%\felix{The above loss function learned a robust embedding space such that the a modified/transformed example stays close to the original embedding, while random examples are far away.}

\paragraph{Encoder Architecture}
For input examples with categorical features, $\mathcal{H}, \mathcal{G}$ are typically constructed with an input layer and a multi-layer perceptron (MLP) on top. The input layer is often a concatenation of normalized dense features and multiple sparse feature embeddings, where the sparse feature embeddings are learnt representations stored in embedding tables (In contrast, the input layers for computer vision and language models directly work on raw inputs). In order to make SSL facilitate the supervised learning task, we share the embedding table of sparse features for both neural networks $\mathcal{H}, \mathcal{G}$. Depending on the technique for data augmentation $(h, g)$, the MLPs of $\mathcal{H}$ and $\mathcal{G}$ could also be fully or partially shared.

\paragraph{Connection with Spread-out Regularization \cite{xu2017spreadout}} In the special case where $(h, g)$ are identical map and $\mathcal{H}, \mathcal{G}$ are the same neural network, loss function in equation \eqref{ssl_loss} is then reduced to $$-N^{-1}\sum_i\log \frac{exp(1/\tau)}{exp(1/\tau) + \sum_{j \ne i} exp(s(\mathbf{z}_i, \mathbf{z}_j)/\tau)}$$which encourages learned representations of different examples to have small cosine similarity. The loss is similar to the spread-out regularization introduced in \cite{xu2017spreadout}, except that the original proposal uses square loss, i.e., $N^{-1} \sum_i \sum_{j \ne i}\langle \mathbf{z}_i, \mathbf{z}_j\rangle^2$, instead of softmax. Spread-out regularization has been proven to improve generalization of large-scale retrieval models. In Section \ref{sec:exp}, we show that by introducing specific data augmentations, using SSL-based regularization can further improve model performance compared to spread-out.

%In this section, we propose a framework for self-supervised learning in the context of recommendation systems. In particular, we leverage a two-tower deep neural net (a.k.a., dual encoders \cite{ref, ref, ref}) model architecture that has been pretty popular for its flexibility and efficiency in computation.

%\subsection{Two-Tower DNN Recommender}
%Figure \ref{fig:two_tower_dnn} show cases the model architecture for a typical two-tower DNN model. On the left, we have a tower to learn query side latent representation (i.e., embeddings) given query information, and similarly the right side tower learns candidate side latent representation given candidate information. The model does a dot product on top to make final predictions between the query and the candidate. Depending on the supervised label and loss functions we pick, this model architecture could be either used for binary / multi-class classification problems, or regression problems.

%This architecture works really well especially for the retrieval stage in recommenders \cite{ref}, or ads targeting for Ads systems \cite{ref}. And it also has been the backbone for many natural language understanding systems  \cite{pre-training} for search retrieval and similar query finding.

%Some background of two-tower DNN recommender, including model-architecture, loss function formulation and it focuses on industrial-scale recommendation problems and many researches have done studying such kind of model \cite{some sir paper}.

\subsection{A Two-stage Data Augmentation} \label{method:ssl_tasks}
%\felix{The most critical question to answer is how to design the transformation. A good transformation and data augmentation should make minimal amount of assumptions on the data such that it can be applicable to a large variety of tasks.}

We introduce the data augmentation, i.e., $h$ and $g$ in Figure \ref{fig:ssl_framework}. Given a set of item features, the key idea is to create two augmented examples by masking part of the information. A good transformation and data augmentation should make minimal amount of assumptions on the data such that it can be generally applicable to a large variety of tasks and models. The idea of masking is inspired by the Masked Language Modeling in BERT \cite{devlin2018bert}. Different from sequential tokens, a general set of features does not have sequential order, and leaves the choice of masking pattern as an open question. We seek to design the masking pattern by exploring feature correlation. We propose Correlated Feature Masking (CFM), tailored to categorical features with awareness of feature correlations.

Before diving into the details of masking, we first present a two-stage augmentation algorithm. Note that without augmentation, the input layer is created by concatenating the embeddings of all categorical features. The two-stage augmentation includes:

\squishlist
\item \textbf{Masking.} Apply a masking pattern on the set of item features. We use a default embedding in the input layer to represent the features that are masked.
\item \textbf{Dropout.} For categorical features with multiple values, we drop out each value with a certain probability. It further reduces input information and increase the hardness of SSL task.
\squishend

The masking step can be interpreted as a special case of dropout with a 100\% dropout rate. One strategy is the complementary masking pattern, that we split the feature set into two exclusive features sets into the two augmented examples.  Specifically, we could randomly split the feature set into two disjoint subsets. We call this method \textbf{Random Feature Masking (RFM)}, and will use it as one of our baselines. We now introduce \textbf{Correlated Feature Masking (CFM)} where we further explore the feature correlations when creating masking patterns.

\paragraph{Mutual Information of Categorical Features.}
If the set of masked features are chosen at random, $(h, g)$ are essentially sampled from $2^k$ different masking patterns over the whole feature set with $k$ features. Different masking patterns would naturally lead to different effects for the SSL task. For instance, the SSL contrastive learning task may exploit the shortcut of highly correlated features between the two augmented examples, making the SSL task too easy. To address this issue, we propose to split features according to feature correlation measured by mutual information. The mutual information of two categorical features is given by
\begin{equation}
MI(V_i, V_j) = \sum_{v_i\in V_i,v_j\in V_j}P(v_i, v_j)\log\frac{P(v_i, v_j)}{P(v_i)P(v_j)},
\end{equation}
where $V_i$, $V_j$ denote their vocab sets. The mutual information for all pairs of features can be pre-computed.

\paragraph{Correlated Feature Masking}
With the pre-computed mutual information, we propose Correlated Feature Masking (CFM) that exploits the feature-dependency patterns for more meaningful SSL tasks. For the set of masked features, $F_m$, we seek to mask highly correlated features together. We do so by first uniformly sample a seed feature $f_{seed}$ from all the available features $F = \{f_1, ..., f_k\}$, and then select the top-n most correlated features $F_{c} = \{ f_{c, 1}, ..., f_{c, n}\}$ according to their mutual information with $f_{seed}$. The final $F_m$ will be the union of the seed and set of correlated features, i.e., $F_m = \{f_{seed}, f_{c, 1}, ..., f_{c, n}\}$. We choose $n = \lfloor k/2 \rfloor$ so that the masked and retained set of features have roughly the same size. We change the seed feature per batch so that the SSL task will learn on various kinds of masking patterns.

%However, not every pair of $h$ and $g$ are equally effective for the SSL task. For instance, $h$ and $g$ with overlapped features sets after the dropout will more likely lead to shortcut in learning the SSL tasks and the probability of $h$ and $g$ has non-overlapping features (bounded by $0.5^k$) in RFM diminishes exponentially with $k$. In one extreme case, if $h$ and $g$ retain the same set of features, the SSL task will be trivial. 

% Google Drawing Locations:
% https://docs.google.com/drawings/d/1Ej0OUhdU3fgkL6f8X1Uav74HefAb4yrA8qiae_U3wyc/edit?resourcekey=0-PNSDpX5pfwUr-O3G47dVhA
\begin{figure*}
    \centering
    \includegraphics[width=0.8\textwidth]{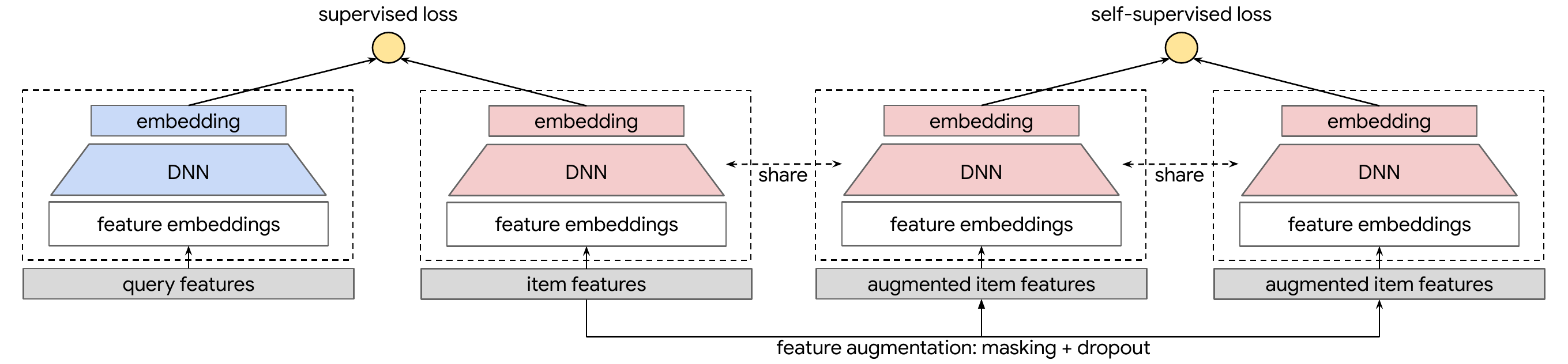}
    \caption{Model architecture: Two-tower model with SSL. In the SSL task, we apply feature masking and dropout on the item features to learn item embeddings. The whole item tower (in red) is shared with the supervised task.}
    \label{fig:feature_dropout_task}
\end{figure*}

\subsection{Multi-task Training} \label{method:ssl_multitask}
To enable SSL learned representations to help improve the learning for the main supervised task such as regression or classification, we leverage a multi-task training strategy where the main supervised task and the auxiliary SSL task are jointly optimized. Precisely, let $\{(q_i, x_i)\}$ be a batch of query-item pairs sampled from the training data distribution $\mathcal{D}_{train}$, and let $\{x_i\}$ be a batch of items sampled from an item distribution $\mathcal{D}_{item}$. Then the joint loss is:
\begin{equation}\label{eq:combined_loss}
\mathcal{L} =  \mathcal{L}_{main}\big(\{(q_i, x_i)\}\big) + \alpha  \cdot \mathcal{L}_{self}\big(\{x_i\}\big),
\end{equation}
where $\mathcal{L}_{main}$ is the loss function for the main task capturing interaction between query and item, and $\alpha$ is the regularization strength.

\paragraph{Heterogeneous Sample Distributions.}
The marginal item distribution from $\mathcal{D}_{train}$ typically follows a power-law. Therefore, using the training item distribution for $\mathcal{L}_{self}$ would cause the learned feature relationship to be biased towards head items. Instead, we sample items uniformly from the corpus for $\mathcal{L}_{self}$. In other words, $\mathcal{D}_{item}$ is the uniform item distribution. In practice, we find using the heterogeneous distributions for main and ssl tasks is critical for SSL to achieve superior performance.

\paragraph{Loss for Main Task.}
There could be many choices for the main loss depending on the objectives. In this paper, we consider the batch softmax loss used in both recommenders \cite{yi2019samplingbias} and NLP \cite{gillick2018endtoend} for optimizing top-k accuracy. In detail, let $\mathbf{q}_i, \mathbf{x}_i$ be the embeddings of query and item examples $(q_i, x_i)$ after being encoded by two neural networks, then for a batch of pairs $\{(q_i, x_i)\}_{i=1}^{N}$ and temperature $\tau$, the batch softmax cross entropy loss is
\begin{equation} \label{main_loss}
\mathcal{L}_{main} = -\frac{1}{N} \sum_{i \in [N]}\log \frac{\exp{(s(\mathbf{q}_i, \mathbf{x}_i)/
\tau)}}{\sum_{j \in [N]} \exp{(s(\mathbf{q}_i,\mathbf{x}_j)/\tau)}}.
\end{equation}

\paragraph{Other Baselines.}
As mentioned in Section \ref{sec:related_work}, we use two-tower DNNs as the baseline model for main task. Two-tower model has the unique property of encoding item features compared to classic matrix factorization (MF) and classification models. While the latter two methods are also applicable to large-scale item retrieval, they only learn item embeddings based on IDs, and thus do not fit in our proposal of using SSL for exploiting item feature relations.

\section{Offline Experiments}\label{sec:exp}
We provide empirical results to demonstrate the effectiveness of our proposed self-supervised framework both in academic public dataset and in actual large-scale recommendation products. The experiments are designed to answer the following research questions.

% \zcheng{Research Questions are supposed to be literally questions with question marks than statements.}
\squishlist
    \item \textbf{RQ1}: Does the proposed SSL Framework improve deep models for recommendations?
    \item \textbf{RQ2}: SSL is designed to improve primary supervised task through introduced SSL task on unlabeled examples. What is the impact of the amount of training data on the improvement from SSL?
    \item \textbf{RQ3}: How do the SSL parameters, i.e., loss multiplier $\alpha$ and dropout rate in data augmentation, affect model quality?
    \item \textbf{RQ4}: How does RFM perform compared to CFM? What is the benefit of leveraging feature correlations in data augmentation?
    %\item \textbf{RQ4}: Do we see a clear trend that SSL helps improve model quality more especially with less supervision?
    %\xinyang{Looks similar to RQ2.}
\squishend
The above questions are addressed in order from Section \ref{sec:impact_rec_quality} - \ref{sec:exp_data_augmentation}.
%\xinyang{Consolidated to 5 questions.}

\subsection{Datasets} \label{sec:datasets}
We conduct experiments on two large-scale datasets that both come with a rich set of item metadata features. We formulate their primary supervised task as an item-to-item recommendation problem to study the effects of SSL on training recommender (in this case, retrieval) models. See Appendix \ref{appendix:data_stats} for details about the statistics of these two datasets.

\textbf{Wikipedia \cite{wikipedia}}: The first dataset focuses on the problem of link prediction between Wikipedia pages. It consists of pairs of pages $(x, y) \in \chi \times \chi$, where $x$ indicates a \textsl{source page}, and $y$ is a \textsl{destination page} linked from $x$. The goal is to predict the set of pages that are likely to be linked to a given source page from the whole corpus of web pages. Each page is represented by a feature vector $x = (x_{id}, x_{ngrams}, x_{cats})$, where all the features are categorical. Here, $x_{id}$ denotes the one-hot encoding of the page URL, $x_{ngrams}$ denotes a bag-of-words representation of the set of n-grams of the page's title, and $x_{cats}$ denotes a bag-of-words representation of the categories that the page belongs to. We partitioned the dataset into training and evaluation using a $(90\%, 10\%)$ split, following the same treatment in \cite{krichene2018efficient} and \cite{yi2019samplingbias}.

\textbf{App-to-App Install (AAI)}: The AAI dataset was collected on the app landing pages from a commercial mobile app store. On a particular app's (seed app) landing page, the app installs (candidate apps) from the section of recommended apps were collected. Each training example represents a pair of seed-candidate pairs denoted as $(x_{seed}, x_{candidate})$ and their metadata features. The goal is to recommend highly similar apps given a seed app. This is also formulated as an item-to-item
recommendation problem via a multi-class classification loss. Note that we only collect positive examples, i.e., $x_{candidate}$ is an installed app from the landing page of $x_{seed}$. All the impressed recommended apps with no installs are all ignored since we consider them more like weak positives instead of negatives for building retrieval models. Each item (app) is represented by a feature vector $\textbf{x}$ with the following features:
\squishlist
    \item \textit{id}: Application id as a one-hot categorical feature.
    
    \item \textit{developer\_name}: Name of the app developer as a one-hot categorical feature. 
    
    \item \textit{categories}: Semantic categories of the app as a multi-hot categorical feature.
    
    \item \textit{title\_unigram}: Uni-grams of the app title as a multi-hot categorical feature.
\squishend
% This is much better now. thanks, Tiansheng!

\subsection{Experiment Setup}

\paragraph{Backbone Network} For the main task that predicts relevant items given the query, we use the two-tower DNN to encode query and items features (see Figure \ref{fig:two_tower_dnn}) as the \textsl{backbone network}. The item-to-item recommendation problem is formalized as a multi-class classification problem, using the batch softmax loss presented in Equation \eqref{main_loss} as the loss function. For discussions of the choice of backbone network, we refer the readers to related sections in Section~\ref{sec:related_work} and Section~\ref{method:ssl_multitask}.

% We choose the two-tower DNN architecture since it uniquely fits two primary considerations: (1) efficiency in recommending from a deep corpus; (2) effectiveness in modeling item-side contextual information. Thus, this simple yet powerful model architecture is dominant in industrial scale recommendation systems \cite{yi2019samplingbias, paul2016}. Note that the proposed framework is itself is architecture agnostic and applicable to networks with sparse categorical features in domains such as pCTR prediction and search ranking, but focus on retrieval models in this work. 

\paragraph{Hyper-parameters} For the backbone two-tower DNN, we search the set of hyper-parameters such as the learning rate, softmax temperature ($\tau$) and model architecture that gives the highest \textit{Recall@50} on the validation set. Note that the training batch size in batch softmax is critical for model quality as it determines the number of negatives used for each positive item. Throughout this section, we use batch sizes $1024$ and $4096$ for Wikipedia and AAI respectively. We also tuned the number of hidden layers, hidden layer sizes and softmax temperature $\tau$ for the baseline models.
For Wikipedia dataset, we use softmax temperature $\tau = 0.07$, and $hidden\_layers$ with sizes $[1024, 128]$. For AAI, we use $\tau=0.06$ and $hidden\_layers$ $[1024, 256]$. Note that the dimension of last hidden layer is also the dimension of final query and item embeddings. All models are trained with Adagrad \cite{adagrad} optimizer with learning rate $0.01$.

We consider two SSL parameters: 1) the SSL loss multiplier $\alpha$ in equation~\eqref{eq:combined_loss}, and 2) the feature dropout rate, denoted as $dr$, in the second phase of data augmentation (see Section \ref{method:ssl_tasks}). For each augmentation method (e.g., CFM, RFM), we conduct grid search of the two parameters by ranges $\alpha = [0.1, 0.3, 1.0, 3.0], dr = [0.1, 0.2, ..., 0.9]$, and report the best result. 

\paragraph{Evaluation} To evaluate the recommendation performance given a seed item, we compute and find the top $K$ items with the highest cosine similarity from the \textbf{whole corpus} and evaluate the quality based on the $K$ retrieved items. Note this is a relatively challenging task, given the sparsity of the dataset and large number of items in the corpus. We adopt popular standard metrics $Recall@K$ and mean average precision ($MAP@K$) to evaluate recommendation performance \cite{He2017}. For each configuration of experiment results, we ran the experiment $5$ times and report the average. 

\subsection{Effectiveness of SSL with Correlated Feature Masking} \label{sec:impact_rec_quality}
To answer \textbf{RQ1}, we first evaluate the impact of SSL on model quality. We focus on using CFM followed by dropout as the data augmentation technique. We will show the superior performance of CFM over other variants in Section~\ref{sec:exp_data_augmentation}.

We consider three baseline methods:
\squishlist
    \item \textit{Baseline}: Vanilla backbone network with the two-tower DNN architecture.
    \item \textit{Feature Dropout (FD)} \cite{Volkovs2017}: Backbone model with random feature dropout on the item tower in the supervised learning task. The feature dropout on item features could be treated as data augmentation. FD does not have the additional SSL regularization compared to our approach.
    \item \textit{Spread-out Regularization (SO)} \cite{xu2017spreadout}: Backbone model with spread-out regularization on the item tower as a regularization. The SO regularization shares similar contrastive loss as that in our SSL framework. However, it applies contrastive learning on original examples without any data augmentation, and is thus different from our approach. 
\squishend
The latter two methods are chosen since they are (1) model-agnostic and scalable for industrial-size recommendation systems; (2) compatible with categorical sparse features for improving generalization. In addition, FD can be viewed as an ablation study to isolate the potential improvement from contrastive learning. Similarly, SO is included to isolate the improvement from feature augmentation.

% the script to remove the std error
% https://colab.corp.google.com/drive/11GDYR8HEEO3qrO2ZW9qwwIxWpRvnVTGN?usp=sharing
\begin{table}[]
    \centering
    \begin{tabular}{c|c|c|c|c}
    \toprule
         \multicolumn{5}{c}{Wikipedia} \\\toprule
         Method & MAP@10 & MAP@50 & Recall@10 & Recall@50  \\\hline
         Baseline& 0.0171 & 0.0229 & 0.0537 & 0.1930 \\\hline
         FD \cite{Volkovs2017}&   0.0172 & 0.0229 & 0.0535 & 0.1912 \\\hline
         SO \cite{xu2017spreadout}&       0.0176 & 0.0235 & 0.0549 & 0.1956 \\\hline
         Our method & \textbf{0.0183} & \textbf{0.0243} & \textbf{0.057} & \textbf{0.2009} \\\toprule
         \multicolumn{5}{c}{AAI} \\\toprule
         Method & MAP@10 & MAP@50 & Recall@10 & Recall@50  \\\hline
         Baseline& 0.1257 & 0.1363 & 0.2793 & 0.4983 \\\hline
         FD  \cite{Volkovs2017}&   0.1278 & 0.1384 & 0.2840 & 0.5058 \\\hline
         SO \cite{xu2017spreadout}&       0.1300 & 0.1406 & 0.2870 & 0.5076 \\\hline
         Our method &       \textbf{0.1413} & \textbf{0.1522} & \textbf{0.3078} & \textbf{0.5355}\\\bottomrule
    \end{tabular}
    \caption{Results on the full Wikipedia and AAI dataset.}
    \vspace{-15pt}
    \label{tab:wiki_results}
\end{table}

We observe that with \textbf{full datasets} (see Table \ref{tab:wiki_results}), CFM consistently performs the best compared with non-SSL regularization techniques. On AAI, CFM out-performs the next best method by 8.69\% relatively and on AAI by 3.98\%. This helps answer \textbf{RQ1} that the proposed SSL framework and tasks indeed improves model performance for recommenders. By comparing CFM with SO, it shows that the data augmentation is critical for the SSL regularization to have better performance. Without any data augmentation, the proposed SSL method is reduced to SO. By comparing CFM and FD, we find the feature augmentation is more effective when applied to the SSL task than to the supervised task as a standard regularization technique. Note that FD, as a well known approach for improving generalization in some cases, applies feature augmentation together with supervised training.

%Our hypothesis is that the value of feature augmentation is better captured via an auxiliary task as in our SSL multitask formulation, where we rely multitask learning to leverage the supervised and self-supervised objective, instead of directly applying on supervised learning as a regularization.

%\xinyang{I feel it would be more clear to only focus on discussing full dataset results here. If some observations are different when data gets sparser, let's discuss it in the next section.} \zcheng{I've broken down the tables into full datasets and sparse datasets. I agree this would be more clear to get the message out}:
    %The worse performance with dropout is perhaps due to the fact that the original baseline model is trained with sufficient training data, and does not really have an over-fitting situation.
    %\item We also see encouraging results when combining both tasks together (FD+FM). On the AAI dataset, FM+FD gives the best performance, showing good complementary between the two SSL tasks. To partially address \textbf{RQ4}, we observe that FD consistently outperforms FM, and we have seen cases that hybrid approach that combines both FD and FM could potentially outperform FD only.
 
\paragraph{Head-tail Analysis.} To understand the gain from SSL, we further break down the overall performance by looking at different item slices by item popularity. The splitting of the \textit{head} and \textit{tail} test set is described in the appendix~\ref{sec_appendix:head_and_tail_data}. Our hypothesis is that SSL generally helps improve the performance for slices without much supervision (e.g., tail items). The results evaluated on the tail and head test sets are reported the results in Table~\ref{tab:wiki_head_and_tail}. We observe that the proposed SSL methods improve the performance for both head and tail item recommendations, with larger gains from the tail items. For instance, in AAI, the CFM improves over $51.5\%$ of the Recall@10 on tail items, while the improvement is $8.57\%$ on head.
 
\begin{figure}
    \small
    \centering
    \includegraphics[width=0.43\columnwidth]{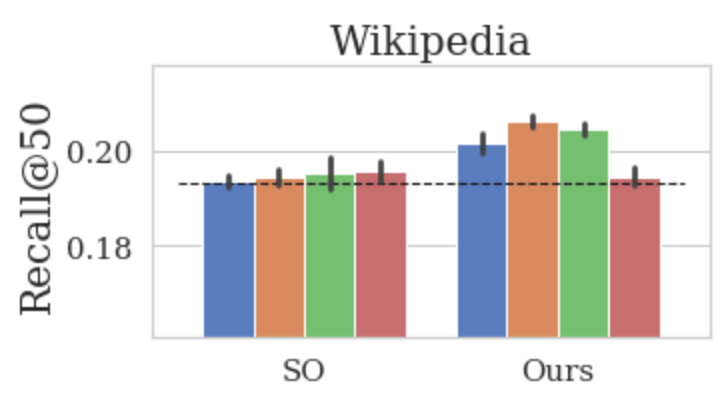}
    \includegraphics[width=0.43\columnwidth]{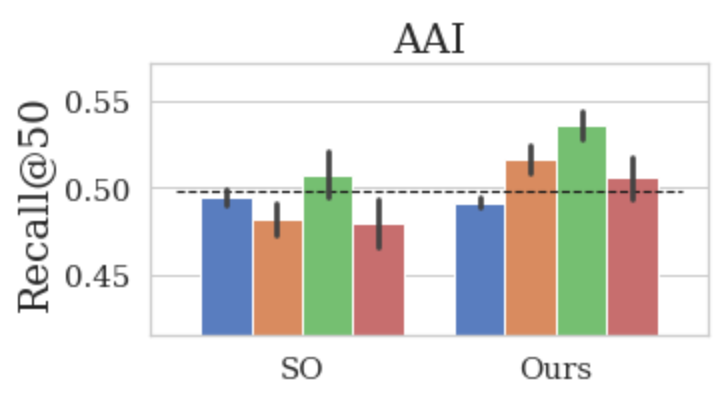}
    \includegraphics[width=0.1\columnwidth]{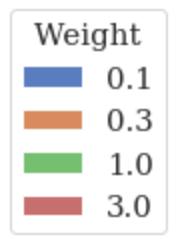}
    \caption{Impact of regularization strength $\alpha$ for SSL-based and spread-out regularization methods. The vertical dash line indicates the baseline model's metric.}
    \label{fig:ssl_weight}
\end{figure}

\begin{figure}
   \small
    \centering
    \includegraphics[width=.48\columnwidth]{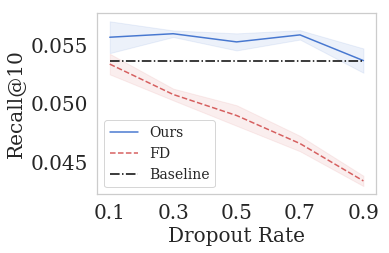}
    \centering
    \includegraphics[width=.48\columnwidth]{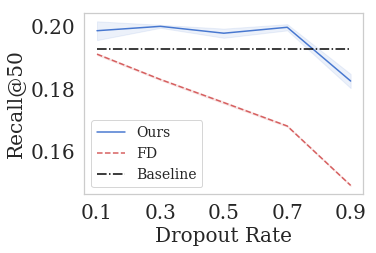}
    \centering
    \includegraphics[width=.48\columnwidth]{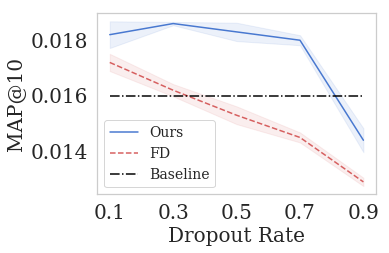}
    \centering
    \includegraphics[width=.48\columnwidth]{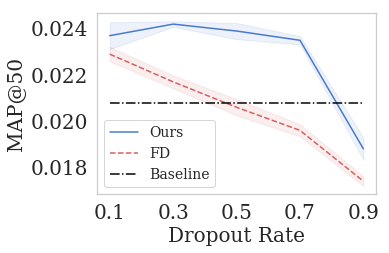}
    \caption{Impact of dropout rate $dr$ for our method and the standard dropout training on the Wikipedia dataset. }\label{fig:fd_wiki}
    \vspace{-10pt}
    %\xinyang{Is it a plan to include similar diagram for AAI?}
    %\tyao{No we don't. The trend is there, but the graph is loaded with confidence interval.}
\end{figure}

\paragraph{Effects of SSL Parameters (\textbf{RQ3}).} Figure \ref{fig:ssl_weight} summarizes the Recall@50 evaluated on the Wikipedia and AAI dataset w.r.t. the regularization strength $\alpha$. It also shows the results of SO which shares the same regularization parameter. We observe that with increasing $\alpha$, the model performance is worse than the baseline model (shown in dash line) after certain threshold. This is expected, since large SSL weight $\alpha$ leads to the multitask loss $\mathcal{L}$ dominated by $\alpha \cdot \mathcal{L}_{self}$ in  equation \eqref{eq:combined_loss}. By further comparing our approach with SO, we show that the SSL based regularization outperforms SO in a wide range of $\alpha$. Figure \ref{fig:fd_wiki} shows the model performance across different dropout rates $dr$. It also shows $DO$ which shares the same parameter. As $dr$ increases, the model performance of $DO$ continues to deteriorate. For most choices of $\alpha$ (except $\alpha = 0.1$), $DO$ is worse than the baseline. For the SSL task with feature dropout, the model performance peaks when $dr = 0.3$ and then deteriorates when we further improve dropout rate. The model starts to under-perform the baseline when $dr$ is too large. This observation aligns with our expectation in the sense that when the rate is too large, the input information becomes too little for to learn meaningful representations through SSL.

\begin{figure}
\centering
\subfloat[Baseline Model]{\label{fig:viz_baseline}
    \centering
    \includegraphics[width=0.65\linewidth]{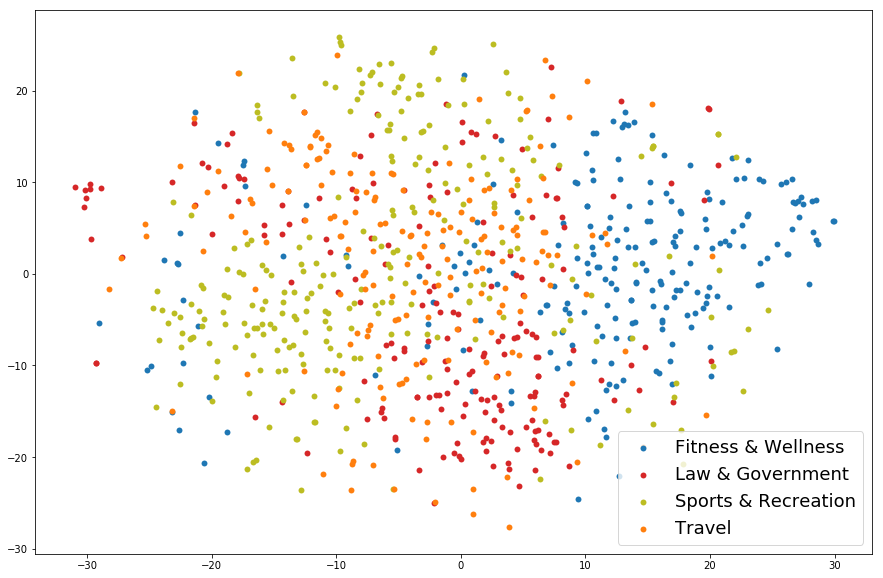}
}
%no space
\hfill
\subfloat[Best SSL Model]{\label{fig:viz_cfd}
    \centering  
    \includegraphics[width=0.65\linewidth]{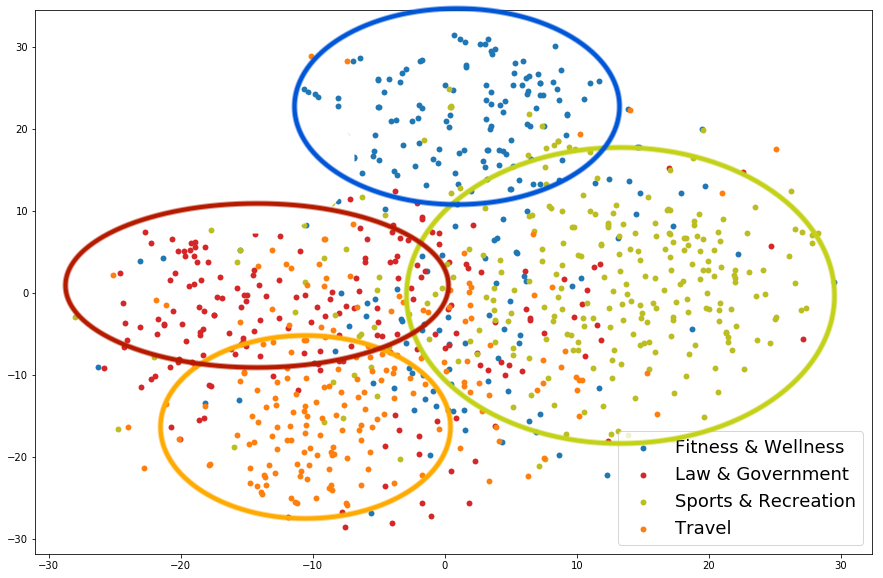}
}
\caption{Comparison of t-SNE plots for app embeddings for baseline, and our best SSL model.}
\label{fig:visualization}
\end{figure}

\paragraph{Visualization of Item Representations.} We visualize the learned app embeddings from the AAI dataset in t-SNE plot. We postpone the detailed setup to Appendix~\ref{sec:visualization}. As shown in in Figure \ref{fig:visualization}, we clearly see that apps embeddings learned with our SSL regularization are better clustered according to their own categories, compared to the counter parts of our baseline, which demonstrates that representations learned through SSL have stronger semantic structures. This partially explains the gain from SSL.

\subsection{Data Sparsity}\label{sec:sparse_exp}
We study the effectiveness of CFM in presence of sparse data to address \textbf{RQ2}. We uniformly down-sampled 10\% of training data and evaluate on the same (full) test dataset. Experiment results are reported in Table \ref{tab:wiki_results_sparse_data}. With increased data sparsity, CFM demonstrates larger improvement for Wikipedia and AAI respectively. In particular, the CFM on the full Wikipedia dataset improves Recall@10 by 6.1\% compared to the baseline, while the relative improvement is 20.6\% on the 10\% dataset. Similar trend is observed for the AAI dataset (10.2\% vs 25.7\% on the down-sampled dataset). It's worth noting that, CFM consistently outperforms FD and the gap is larger as data becomes sparser. This demonstrates that having dropout for data augmentation in SSL tasks is more effective than directly applying dropout in supervised task. 

As a summary, these findings answer research questions raised in \textbf{RQ2} that the proposed SSL framework improves model performance more with even less supervision.
%One hypothesis is that the AAI dataset is even more skewed towards popular items as illustrated in Figure \ref{fig:dataset}. In addition, it also includes lots of duplicated pairs of $(query, item)$, where the pairs in Wikipedia are all unique. When we down-sample the AAI dataset to 10\%, many of the $(query, item)$ pairs are still preserved due to duplication, so the information is still well kept. When the data is down-sampled to 1\%, the remaining data is truly sparse of the original supervised information, and thus leads to more obvious improvements from SSL. 

\begin{table}[]
    \centering
    \begin{tabular}{c|c|c|c|c}
    \toprule
         \multicolumn{5}{c}{10\% Wikipedia Dataset}\\ \toprule
                  Method & MAP@10 & MAP@50 & Recall@10 & Recall@50   \\\hline
         Baseline& 0.0077 & 0.0105 & 0.0237 & 0.0924 \\\hline
         FD \cite{Volkovs2017}& 0.0089 &0.0120 & 0.0272 &0.1046 \\\hline
         SO \cite{xu2017spreadout}&       0.0083 & 0.0112 & 0.0254 & 0.0978 \\\hline
         Our method &       \textbf{0.0093}& \textbf{0.0126 }& \textbf{0.0286 }& \textbf{0.1093 }\\
         \toprule
         \multicolumn{5}{c}{10\% AAI Dataset} \\
         \toprule
         Method & MAP@10 & MAP@50 & Recall@10 & Recall@50  \\\hline
         Baseline& 0.1112 & 0.1194 & 0.2406 & 0.4068 \\\hline
         FD  \cite{Volkovs2017}&   0.1217 & 0.1302 & 0.2611 & 0.4324 \\\hline
         SO \cite{xu2017spreadout}&       0.1220 & 0.1308 & 0.2632 & 0.4400 \\\hline
         Our method &       \textbf{0.1409} & \textbf{0.1507} & \textbf{0.3024} & \textbf{0.5014}\\
         \bottomrule
    \end{tabular}
    \caption{Experiment results trained on the sparse (10\% down-sampled) Wikipedia and AAI datasets.}
    \vspace{-10pt}
    \label{tab:wiki_results_sparse_data}
\end{table}

%\subsection{Item Slice Analysis}\label{sec:tail_rec}
%We further break down the overall performance by looking at different groundtruth item slices distinguished by item popularity, to further understand where does the gains come from. The splitting of the \textit{head} and \textit{tail} test set is detailed in Section~\ref{sec_appendix:head_and_tail_data}. Our hypothesis is that SSL generally helps improve the performance for slices without much supervision (e.g., tail items). The results evaluated on the tail and head test sets are reported the results in Table~\ref{tab:wiki_head_and_tail} and Table~\ref{tab:aai_head_and_tail}.

%We observe that the proposed SSL methods improve the performance for both head and tail item recommendations, with larger gains from the tail items. For Wikipedia, CFM improves Recall@10 by $11.0\%$ relatively for tail items, compared to $1.48\%$ on the head items. For AAI, the gap is even larger -- the CFM improves over $51.5\%$ of the Recall@10 on tail items. These results show that the proposed SSL technique is effective in alleviating the label sparsity in the long-tail item distribution, and significantly improves the representation learned for tail items.

%Note for AAI dataset, since labels were collected from actual user engagement, the ground-truth may suffer from lack of implicit feedback and bias from existing system. We also report live experiment results in Section~\ref{sec:online_exp} to further demonstrate the effectiveness of the proposed SSL approaches in a web-scale system.

\begin{table}[t!]
\centering
\begin{tabular}{c|c|c|c|c}
\toprule
\multicolumn{5}{c}{Wikipedia}\\
\toprule
Method  & \multicolumn{2}{c|}{Tail} & \multicolumn{2}{c}{Head}
\\ \cline{2-5}
          & Recall@10 & Recall@50 & Recall@10 & Recall@50 \\\hline
Baseline & 0.0472 & 0.1621 & 0.0610 & 0.2273 \\\hline
FD & 0.0474 & 0.1638 & 0.0593 & 0.2212\\\hline
SO & 0.0481 & 0.1644 & 0.0606 & 0.2268 \\\hline
Our method & \textbf{0.0524} & \textbf{0.1749} & \textbf{0.0619} & \textbf{0.2283}\\
\toprule
\multicolumn{5}{c}{AAI}\\
\toprule
Baseline & 0.0475 & 0.2333 & 0.2846 & 0.4993 \\\hline
FD & \textbf{0.0727} & 0.2743 & 0.2849 & 0.5069 \\\hline
SO & 0.0661 & 0.2602 & 0.2879 & 0.5086 \\\hline
Our method & 0.0720 & \textbf{0.2906} & \textbf{0.309} & \textbf{0.537}
\\ \bottomrule
\end{tabular}
\caption{Results of Wikipedia and AAI on tail and head item slices.}\label{tab:wiki_head_and_tail}
\vspace{-10pt}
\end{table}

%\subsection{Analysis on Hyperparameters}
%\subsubsection{Effect of SSL weight $\alpha$} \label{sec:ssl_weights}

%To address \textbf{RQ3}, we study the effect of SSL regularization strength $\alpha$ presented in equation \eqref{ssl_loss}. We perform a line search of $\alpha$ in $\{0.1, 0.3, 1.0, 3.0\}$ to study the SSL tasks' sensitivity to $\alpha$. \xinyang{This was mentioned before.}

\subsection{Comparison of Different Data Augmentations}\label{sec:exp_data_augmentation}
In this section, we compare several feature augmentation alternatives with CFM to answer \textbf{RQ4} by studying: 1) the benefit of exploiting feature correlation in masking and 2) benefit of using dropout as part of augmentation. In specific, we consider the following alternatives:

\squishlist
    \item \textit{$RFM$}: Random Feature Masking. In this method, random set of features are masked, instead of guided by mutual information in CFM.
    \item \textit{$RFM_{no\_compl}$}: Random Feature Masking with no Complementary sets of features applied. In this method, 2 independent sets of features are masked at random, instead of a complementary pair of masks in CFM.
    \item \textit{$CFM_{no\_dropout}$}: Correlated Feature Masking with No Dropout applied. In other words, only apply correlated masking as the augmentation in the SSL task.
    \item \textit{$NoMasking$}: Correlated Feature Masking but skipping the masking phrase in augmentation. In other words, we only apply dropout to features as the augmentation.
\squishend

We apply these feature augmentation functions in the SSL framework and report the results on AAI dataset in Table~\ref{tab:ablation_result}. 

First, we observe that all the variants are worse than CFM, but still out-perform the baseline model. In particular, we see having mutual information in selecting the masking set is critical to model improvement, since we see the biggest performance drop is from RFM where masking set is selected at random. By comparing CFM with results from the two methods ($RFM_{no\_compl}$ and $NoMasking$) that allow feature overlap between the two augmented examples via independent dropout, we see the contrastive learning task is more helpful with complementary information that potentially avoids shortcut in learning. Finally, by comparing $CFM_{no\_dropout}$ and CFM, we see the second phrase of randomly dropping out feature values also helps, which could be potentially explained by introducing more feature variants in the SSL task.

\begin{table}[]
    \small
    \centering
    \begin{tabular}{c|c|c|c|c}
    \toprule
    \multicolumn{5}{c}{Comparison of Multiple Data Augmentations} \\\toprule
                  Method & MAP@10 & MAP@50 & Recall@10 & Recall@50   \\\hline
         Baseline& 0.1257 & 0.1363 & 0.2793 & 0.4983 \\\hline\hline
         CFM &       \textbf{0.1413} & \textbf{0.1522} & \textbf{0.3078} & \textbf{0.5355} \\\hline\hline
         RFM &       0.1281 & 0.1389 & 0.2851 & 0.5104\\\hline
         $RFM_{no\_compl}$ &       0.1363 & 0.1472 & 0.3007 &	0.5290\\\hline
         $CFM_{no\_dropout}$ &       0.1309 & 0.1417 & 0.2898 &	0.5150\\\hline
         $NoMasking$ &       0.1303 & 0.1408 & 0.2868 &	0.5053\\\bottomrule
    \end{tabular}
    \caption{Results of CFM and other data augmentation techniques on the AAI dataset.}
    \vspace{-10pt}
    \label{tab:ablation_result}
\end{table}

\section{Live Experiment}\label{sec:online_exp}
In this section, we describe how we apply our proposed SSL framework to a web-scale commercial app recommender system. Specifically, given an app as the query, the system identifies similar apps given the query. One of the models surfacing this recommendation is trained on the AAI dataset as described in Section \ref{sec:datasets}, with the same backbone network structure as the two-tower DNN structure in Figure ~\ref{fig:two_tower_dnn} (with modifications). For a natural extension of the offline experiments conducted on Section~\ref{sec:impact_rec_quality} for the AAI experiments, we conducted an A/B experiment for investigating the synergy of deploying the best SSL-based model online. While we already presented improved offline metrics on this dataset, in many real-world systems, offline studies might not align with live impact due to 1) lack of implicit feedback, since the offline evaluation data is collected via user engagement history based on the production system; 2) failing to capture product's multiple objectives optimization goal, where it's very likely that recommending more engaging apps hurts other business goals. Therefore, this experiment is critical in demonstrating the effectiveness of the proposed framework in real-world settings.

\begin{figure}
\subfloat[Global]{\label{fig:rasta_rev}
    \centering  
    \includegraphics[width=0.48\linewidth]{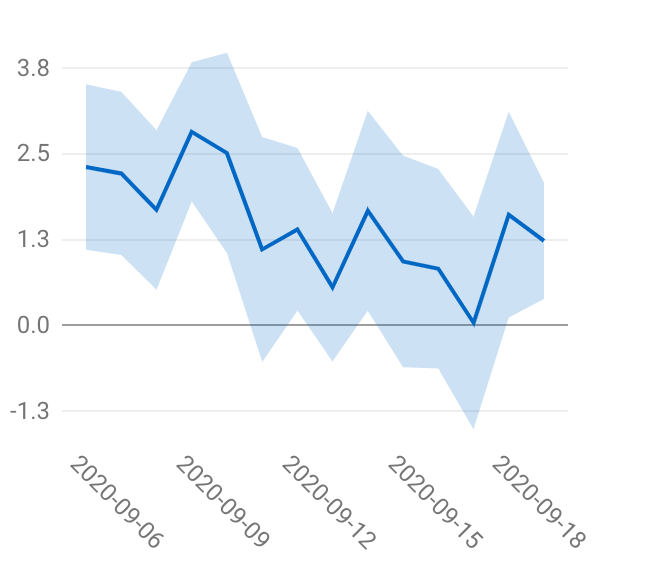}
}%
\hfill
\subfloat[Localized]{\label{fig:rasta_rev_br}
    \centering  
    \includegraphics[width=0.48\linewidth]{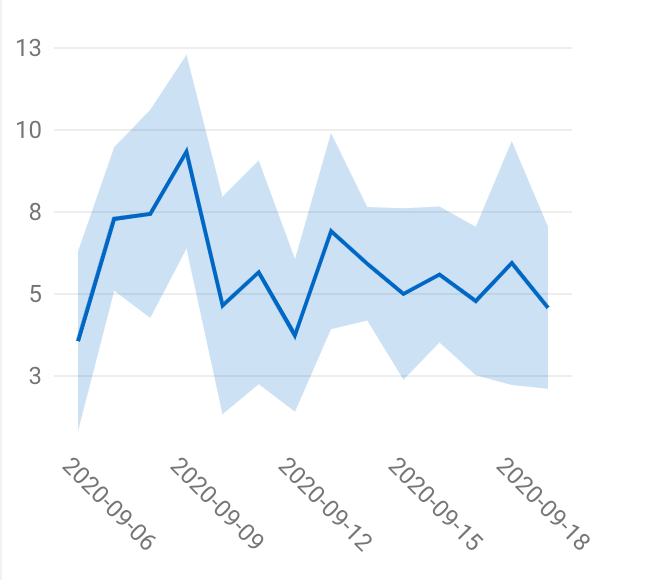}
}
\caption{Top Business Metric Improvement Percentage (y-axis) over Days (x-axis) in Online Experiments: (a) improvement globally; (b) improvement on a localized market.}
\vspace{-15pt}
\end{figure}

\begin{figure}
\subfloat[All Apps]{\label{fig:rasta_install}
    \centering  
    \includegraphics[width=0.48\linewidth]{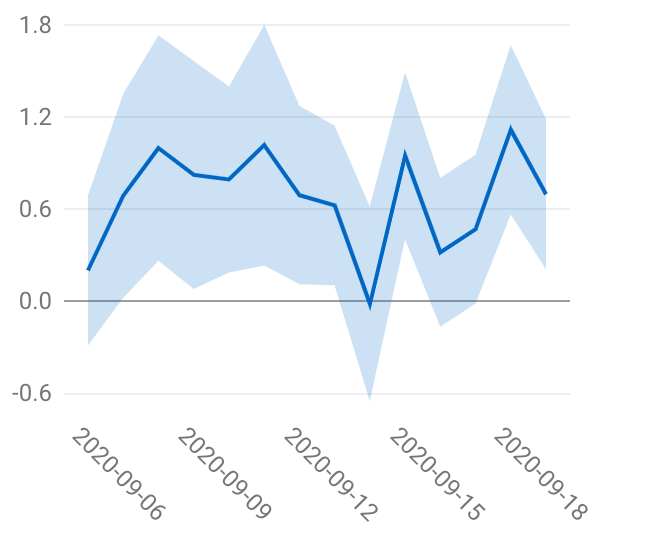}
}%
\hfill
\subfloat[Fresh Apps]{\label{fig:rasta_fresh_install}
    \centering  
    \includegraphics[width=0.48\linewidth]{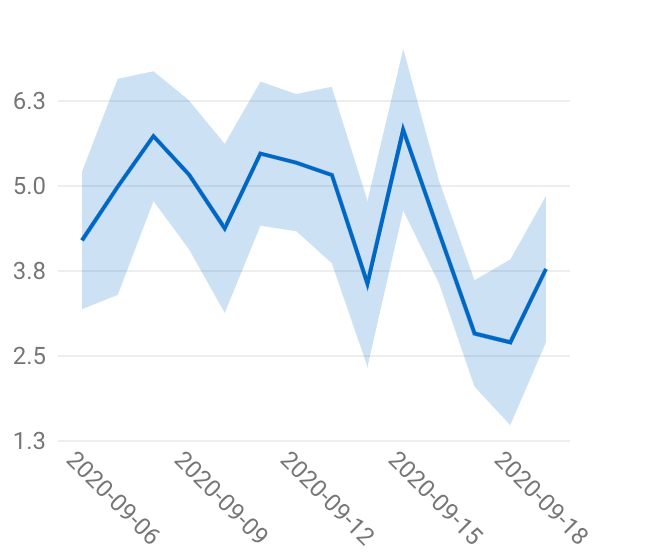}
}
\caption{Top User Engagement Improvement Percentage (y-axis) over Days (x-axis) in Online Experiments: (a) improvement on all apps; (b) improvement on fresh apps.}
\vspace{-15pt}
\end{figure}

%% Rasta: https://rasta.corp.google.com/#/metrics?label=_:udCYArsfY9zOVRzhT3nMvlOEHcE
%% Freshness Rasta: https://rasta.corp.google.com/#/metrics?label=_:ZAtQacZV6OXfV0XaY0nKlQ0EmQc
In our live A/B testing, we add the best performing SSL task with the same set of hyper-parameters on top of the existing well-tuned two-tower DNN model used in production. In a time frame of 14 days, the model improved the overall business metrics significantly, with $+0.67\%$ increase in key user engagement (Figure~\ref{fig:rasta_install}) and $+1.5\%$ increase in top business metric (Figure~\ref{fig:rasta_rev}). To echo the study on the \textit{Head-tail Analysis} in Section~\ref{sec:impact_rec_quality} and the data sparsity analysis in  Section~\ref{sec:sparse_exp}, we see significant improvements on two slices: 1) cold-starting for fresh apps: the model improves $+4.5\%$ on user engagement for fresh apps (Figure~\ref{fig:rasta_fresh_install}); and (2) international countries that have sparser training data compared to major markets: we see significant $+5.47\%$ top business metric gains (Figure~\ref{fig:rasta_rev_br} right). Again, both of these results verify our hypothesis that our SSL framework indeed significantly improves model performance for slices without much supervision. Given the results, the SSL empowered model was successfully launched in the current production system.


\section{Conclusion}

In this paper, we proposed a model architecture agnostic self-supervised learning (SSL) framework for large-scale neural recommender models. Within the SSL framework, we also introduced a novel data augmentation applicable to heterogeneous categorical features and showed its superior performance over other variants.

For future works, we plan to investigate how different training schemes impact the model quality. One direction is to first pre-train on SSL task to learn query and item representations and fine-tune on primary supervised tasks. Alternatively, it would be interesting to extend the technique for deep models in application domains such as search ranking or pCTR prediction.
%\newpage

\bibliographystyle{ACM-Reference-Format}
\bibliography{acmart}


\begin{thebibliography}{43}

%%% the Web address.

\ifx \showCODEN    \undefined \def \showCODEN     #1{\unskip}     \fi
\ifx \showDOI      \undefined \def \showDOI       #1{#1}\fi
\ifx \showISBNx    \undefined \def \showISBNx     #1{\unskip}     \fi
\ifx \showISBNxiii \undefined \def \showISBNxiii  #1{\unskip}     \fi
\ifx \showISSN     \undefined \def \showISSN      #1{\unskip}     \fi
\ifx \showLCCN     \undefined \def \showLCCN      #1{\unskip}     \fi
\ifx \shownote     \undefined \def \shownote      #1{#1}          \fi
\ifx \showarticletitle \undefined \def \showarticletitle #1{#1}   \fi
\ifx \showURL      \undefined \def \showURL       {\relax}        \fi
% invisible to TeX
\providecommand\bibfield[2]{#2}
\providecommand\bibinfo[2]{#2}
\providecommand\natexlab[1]{#1}
\providecommand\showeprint[2][]{arXiv:#2}

\bibitem[\protect\citeauthoryear{Beutel, Chi, Cheng, Pham, and Anderson}{Beutel
  et~al\mbox{.}}{[n.d.]}]%
        {focused_learning_2017}
\bibfield{author}{\bibinfo{person}{Alex Beutel}, \bibinfo{person}{Ed~H. Chi},
  \bibinfo{person}{Zhiyuan Cheng}, \bibinfo{person}{Hubert Pham}, {and}
  \bibinfo{person}{John Anderson}.} \bibinfo{year}{[n.d.]}\natexlab{}.
\newblock \showarticletitle{Beyond Globally Optimal: Focused Learning for
  Improved Recommendations}. In \bibinfo{booktitle}{\emph{WWW 2017}}.
\newblock


\bibitem[\protect\citeauthoryear{{Beyer}, {Zhai}, {Oliver}, and
  {Kolesnikov}}{{Beyer} et~al\mbox{.}}{[n.d.]}]%
        {s4l}
\bibfield{author}{\bibinfo{person}{L. {Beyer}}, \bibinfo{person}{X. {Zhai}},
  \bibinfo{person}{A. {Oliver}}, {and} \bibinfo{person}{A. {Kolesnikov}}.}
  \bibinfo{year}{[n.d.]}\natexlab{}.
\newblock \showarticletitle{S4L: Self-Supervised Semi-Supervised Learning}. In
  \bibinfo{booktitle}{\emph{ICCV 2019}}.
\newblock


\bibitem[\protect\citeauthoryear{Chang, Yu, Chang, Yang, and Kumar}{Chang
  et~al\mbox{.}}{[n.d.]}]%
        {pre-training}
\bibfield{author}{\bibinfo{person}{Wei-Cheng Chang}, \bibinfo{person}{Felix~X.
  Yu}, \bibinfo{person}{Yin-Wen Chang}, \bibinfo{person}{Yiming Yang}, {and}
  \bibinfo{person}{Sanjiv Kumar}.} \bibinfo{year}{[n.d.]}\natexlab{}.
\newblock \showarticletitle{Pre-training Tasks for Embedding-based Large-scale
  Retrieval}. In \bibinfo{booktitle}{\emph{ICLR 2020}}.
\newblock


\bibitem[\protect\citeauthoryear{Chen and Guestrin}{Chen and
  Guestrin}{[n.d.]}]%
        {xgboost}
\bibfield{author}{\bibinfo{person}{Tianqi Chen} {and} \bibinfo{person}{Carlos
  Guestrin}.} \bibinfo{year}{[n.d.]}\natexlab{}.
\newblock \showarticletitle{XGBoost: A Scalable Tree Boosting System}. In
  \bibinfo{booktitle}{\emph{KDD 2016}}.
\newblock


\bibitem[\protect\citeauthoryear{Chen, Kornblith, Norouzi, and Hinton}{Chen
  et~al\mbox{.}}{2020a}]%
        {chen2020simple}
\bibfield{author}{\bibinfo{person}{Ting Chen}, \bibinfo{person}{Simon
  Kornblith}, \bibinfo{person}{Mohammad Norouzi}, {and}
  \bibinfo{person}{Geoffrey~E. Hinton}.} \bibinfo{year}{2020}\natexlab{a}.
\newblock \bibinfo{title}{A Simple Framework for Contrastive Learning of Visual
  Representations}.
\newblock
\newblock
\urldef\tempurl%
\url{https://arxiv.org/abs/2002.05709}
\showURL{%
\tempurl}


\bibitem[\protect\citeauthoryear{Chen, Kornblith, Swersky, Norouzi, and
  Hinton}{Chen et~al\mbox{.}}{2020b}]%
        {chen2020big}
\bibfield{author}{\bibinfo{person}{Ting Chen}, \bibinfo{person}{Simon
  Kornblith}, \bibinfo{person}{Kevin Swersky}, \bibinfo{person}{Mohammad
  Norouzi}, {and} \bibinfo{person}{Geoffrey Hinton}.}
  \bibinfo{year}{2020}\natexlab{b}.
\newblock \showarticletitle{Big Self-Supervised Models are Strong
  Semi-Supervised Learners}.
\newblock \bibinfo{journal}{\emph{arXiv preprint arXiv:2006.10029}}
  (\bibinfo{year}{2020}).
\newblock


\bibitem[\protect\citeauthoryear{Cheng, Koc, Harmsen, Shaked, Chandra, Aradhye,
  Anderson, Corrado, Chai, Ispir, Anil, Haque, Hong, Jain, Liu, and Shah}{Cheng
  et~al\mbox{.}}{[n.d.]}]%
        {heng16}
\bibfield{author}{\bibinfo{person}{Heng-Tze Cheng}, \bibinfo{person}{Levent
  Koc}, \bibinfo{person}{Jeremiah Harmsen}, \bibinfo{person}{Tal Shaked},
  \bibinfo{person}{Tushar Chandra}, \bibinfo{person}{Hrishi Aradhye},
  \bibinfo{person}{Glen Anderson}, \bibinfo{person}{Greg Corrado},
  \bibinfo{person}{Wei Chai}, \bibinfo{person}{Mustafa Ispir},
  \bibinfo{person}{Rohan Anil}, \bibinfo{person}{Zakaria Haque},
  \bibinfo{person}{Lichan Hong}, \bibinfo{person}{Vihan Jain},
  \bibinfo{person}{Xiaobing Liu}, {and} \bibinfo{person}{Hemal Shah}.}
  \bibinfo{year}{[n.d.]}\natexlab{}.
\newblock \showarticletitle{Wide \& Deep Learning for Recommender Systems}
  \emph{(\bibinfo{series}{DLRS 2016})}.
\newblock


\bibitem[\protect\citeauthoryear{Christakopoulou and Karypis}{Christakopoulou
  and Karypis}{[n.d.]}]%
        {chris_kdd_2018}
\bibfield{author}{\bibinfo{person}{Evangelia Christakopoulou} {and}
  \bibinfo{person}{George Karypis}.} \bibinfo{year}{[n.d.]}\natexlab{}.
\newblock \showarticletitle{Local Latent Space Models for Top-N
  Recommendation}.
\newblock


\bibitem[\protect\citeauthoryear{Cohen and Lewis}{Cohen and Lewis}{[n.d.]}]%
        {Cohen1997}
\bibfield{author}{\bibinfo{person}{Edith Cohen} {and} \bibinfo{person}{David~D.
  Lewis}.} \bibinfo{year}{[n.d.]}\natexlab{}.
\newblock \showarticletitle{Approximating Matrix Multiplication for Pattern
  Recognition Tasks}. In \bibinfo{booktitle}{\emph{SODA 1997}}.
\newblock


\bibitem[\protect\citeauthoryear{Covington, Adams, and Sargin}{Covington
  et~al\mbox{.}}{[n.d.]}]%
        {paul2016}
\bibfield{author}{\bibinfo{person}{Paul Covington}, \bibinfo{person}{Jay
  Adams}, {and} \bibinfo{person}{Emre Sargin}.}
  \bibinfo{year}{[n.d.]}\natexlab{}.
\newblock \showarticletitle{Deep Neural Networks for YouTube Recommendations}.
  In \bibinfo{booktitle}{\emph{RecSys 2016}}.
\newblock


\bibitem[\protect\citeauthoryear{Dacrema, Cremonesi, and Jannach}{Dacrema
  et~al\mbox{.}}{[n.d.]}]%
        {dacrema2019really}
\bibfield{author}{\bibinfo{person}{Maurizio~Ferrari Dacrema},
  \bibinfo{person}{Paolo Cremonesi}, {and} \bibinfo{person}{Dietmar Jannach}.}
  \bibinfo{year}{[n.d.]}\natexlab{}.
\newblock \showarticletitle{Are We Really Making Much Progress? A Worrying
  Analysis of Recent Neural Recommendation Approaches}. In
  \bibinfo{booktitle}{\emph{RecSys 2019}}.
\newblock


\bibitem[\protect\citeauthoryear{Devlin, Chang, Lee, and Toutanova}{Devlin
  et~al\mbox{.}}{[n.d.]}]%
        {devlin2018bert}
\bibfield{author}{\bibinfo{person}{Jacob Devlin}, \bibinfo{person}{Ming{-}Wei
  Chang}, \bibinfo{person}{Kenton Lee}, {and} \bibinfo{person}{Kristina
  Toutanova}.} \bibinfo{year}{[n.d.]}\natexlab{}.
\newblock \showarticletitle{{BERT:} Pre-training of Deep Bidirectional
  Transformers for Language Understanding}. In
  \bibinfo{booktitle}{\emph{NAACL-HLT 2019}}.
\newblock


\bibitem[\protect\citeauthoryear{Duchi, Hazan, and Singer}{Duchi
  et~al\mbox{.}}{2011}]%
        {adagrad}
\bibfield{author}{\bibinfo{person}{John Duchi}, \bibinfo{person}{Elad Hazan},
  {and} \bibinfo{person}{Yoram Singer}.} \bibinfo{year}{2011}\natexlab{}.
\newblock \showarticletitle{Adaptive Subgradient Methods for Online Learning
  and Stochastic Optimization}.
\newblock \bibinfo{journal}{\emph{J. Mach. Learn. Res.}} \bibinfo{volume}{12},
  \bibinfo{number}{null} (\bibinfo{date}{July} \bibinfo{year}{2011}),
  \bibinfo{pages}{2121–2159}.
\newblock
\showISSN{1532-4435}


\bibitem[\protect\citeauthoryear{Foundation}{Foundation}{[n.d.]}]%
        {wikipedia}
\bibfield{author}{\bibinfo{person}{Wikimedia Foundation}.}
  \bibinfo{year}{[n.d.]}\natexlab{}.
\newblock \showarticletitle{Wikimedia}.
\newblock
\urldef\tempurl%
\url{https://dumps.wikimedia.org/}
\showURL{%
\tempurl}


\bibitem[\protect\citeauthoryear{Gidaris, Singh, and Komodakis}{Gidaris
  et~al\mbox{.}}{[n.d.]}]%
        {gidaris2018unsupervised}
\bibfield{author}{\bibinfo{person}{Spyros Gidaris}, \bibinfo{person}{Praveer
  Singh}, {and} \bibinfo{person}{Nikos Komodakis}.}
  \bibinfo{year}{[n.d.]}\natexlab{}.
\newblock \showarticletitle{Unsupervised Representation Learning by Predicting
  Image Rotations}. In \bibinfo{booktitle}{\emph{ICLR 2018}}.
\newblock


\bibitem[\protect\citeauthoryear{Gillick, Presta, and Tomar}{Gillick
  et~al\mbox{.}}{2018}]%
        {gillick2018endtoend}
\bibfield{author}{\bibinfo{person}{Daniel Gillick}, \bibinfo{person}{Alessandro
  Presta}, {and} \bibinfo{person}{Gaurav~Singh Tomar}.}
  \bibinfo{year}{2018}\natexlab{}.
\newblock \showarticletitle{End-to-End Retrieval in Continuous Space}.
\newblock \bibinfo{journal}{\emph{CoRR}}  \bibinfo{volume}{abs/1811.08008}
  (\bibinfo{year}{2018}).
\newblock
\urldef\tempurl%
\url{http://arxiv.org/abs/1811.08008}
\showURL{%
\tempurl}


\bibitem[\protect\citeauthoryear{Guo, Mousavi, Wu, Holtmann-Rice, Kale, Reddi,
  and Kumar}{Guo et~al\mbox{.}}{2019}]%
        {guo2019glass}
\bibfield{author}{\bibinfo{person}{Chuan Guo}, \bibinfo{person}{Ali Mousavi},
  \bibinfo{person}{Xiang Wu}, \bibinfo{person}{Daniel~N Holtmann-Rice},
  \bibinfo{person}{Satyen Kale}, \bibinfo{person}{Sashank Reddi}, {and}
  \bibinfo{person}{Sanjiv Kumar}.} \bibinfo{year}{2019}\natexlab{}.
\newblock \showarticletitle{Breaking the Glass Ceiling for Embedding-Based
  Classifiers for Large Output Spaces}.
\newblock In \bibinfo{booktitle}{\emph{Neurips}},
  \bibfield{editor}{\bibinfo{person}{H.~Wallach},
  \bibinfo{person}{H.~Larochelle}, \bibinfo{person}{A.~Beygelzimer},
  \bibinfo{person}{F.~d\textquotesingle Alch\'{e}-Buc},
  \bibinfo{person}{E.~Fox}, {and} \bibinfo{person}{R.~Garnett}} (Eds.).
\newblock


\bibitem[\protect\citeauthoryear{He, Liao, Zhang, Nie, Hu, and Chua}{He
  et~al\mbox{.}}{[n.d.]}]%
        {He2017}
\bibfield{author}{\bibinfo{person}{Xiangnan He}, \bibinfo{person}{Lizi Liao},
  \bibinfo{person}{Hanwang Zhang}, \bibinfo{person}{Liqiang Nie},
  \bibinfo{person}{Xia Hu}, {and} \bibinfo{person}{Tat-Seng Chua}.}
  \bibinfo{year}{[n.d.]}\natexlab{}.
\newblock \showarticletitle{Neural Collaborative Filtering}. In
  \bibinfo{booktitle}{\emph{WWW 2017}}.
\newblock


\bibitem[\protect\citeauthoryear{He, Pan, Jin, Xu, Liu, Xu, Shi, Atallah,
  Herbrich, Bowers, and Candela}{He et~al\mbox{.}}{2014}]%
        {fb_adkdd_14}
\bibfield{author}{\bibinfo{person}{Xinran He}, \bibinfo{person}{Junfeng Pan},
  \bibinfo{person}{Ou Jin}, \bibinfo{person}{Tianbing Xu}, \bibinfo{person}{Bo
  Liu}, \bibinfo{person}{Tao Xu}, \bibinfo{person}{Yanxin Shi},
  \bibinfo{person}{Antoine Atallah}, \bibinfo{person}{Ralf Herbrich},
  \bibinfo{person}{Stuart Bowers}, {and} \bibinfo{person}{Joaquin Qui\~{n}onero
  Candela}.} \bibinfo{year}{2014}\natexlab{}.
\newblock \showarticletitle{Practical Lessons from Predicting Clicks on Ads at
  Facebook}. In \bibinfo{booktitle}{\emph{Proceedings of the Eighth
  International Workshop on Data Mining for Online Advertising}}.
\newblock


\bibitem[\protect\citeauthoryear{Kolesnikov, Zhai, and Beyer}{Kolesnikov
  et~al\mbox{.}}{[n.d.]}]%
        {kolesnikov2019revisiting}
\bibfield{author}{\bibinfo{person}{Alexander Kolesnikov},
  \bibinfo{person}{Xiaohua Zhai}, {and} \bibinfo{person}{Lucas Beyer}.}
  \bibinfo{year}{[n.d.]}\natexlab{}.
\newblock \showarticletitle{Revisiting Self-Supervised Visual Representation
  Learning}. In \bibinfo{booktitle}{\emph{CVPR 2019}}.
\newblock


\bibitem[\protect\citeauthoryear{Koren, Bell, and Volinsky}{Koren
  et~al\mbox{.}}{2009}]%
        {matrix_factorization}
\bibfield{author}{\bibinfo{person}{Yehuda Koren}, \bibinfo{person}{Robert
  Bell}, {and} \bibinfo{person}{Chris Volinsky}.}
  \bibinfo{year}{2009}\natexlab{}.
\newblock \showarticletitle{Matrix Factorization Techniques for Recommender
  Systems}.
\newblock \bibinfo{journal}{\emph{Computer}} \bibinfo{volume}{42},
  \bibinfo{number}{8} (\bibinfo{date}{Aug.} \bibinfo{year}{2009}),
  \bibinfo{pages}{30–37}.
\newblock
\showISSN{0018-9162}


\bibitem[\protect\citeauthoryear{Koren and Bell}{Koren and Bell}{2015}]%
        {advanced_cf}
\bibfield{author}{\bibinfo{person}{Yehuda Koren} {and}
  \bibinfo{person}{Robert~M. Bell}.} \bibinfo{year}{2015}\natexlab{}.
\newblock \bibinfo{booktitle}{\emph{Advances in Collaborative Filtering}}.
\newblock \bibinfo{publisher}{Springer}, \bibinfo{pages}{77--118}.
\newblock


\bibitem[\protect\citeauthoryear{Krichene, Mayoraz, Rendle, Zhang, Yi, Hong,
  Chi, and Anderson}{Krichene et~al\mbox{.}}{[n.d.]}]%
        {krichene2018efficient}
\bibfield{author}{\bibinfo{person}{Walid Krichene}, \bibinfo{person}{Nicolas
  Mayoraz}, \bibinfo{person}{Steffen Rendle}, \bibinfo{person}{Li Zhang},
  \bibinfo{person}{Xinyang Yi}, \bibinfo{person}{Lichan Hong},
  \bibinfo{person}{Ed Chi}, {and} \bibinfo{person}{John Anderson}.}
  \bibinfo{year}{[n.d.]}\natexlab{}.
\newblock \showarticletitle{Efficient Training on Very Large Corpora via
  Gramian Estimation}. In \bibinfo{booktitle}{\emph{ICLR 2019}}.
\newblock


\bibitem[\protect\citeauthoryear{Lan, Chen, Goodman, Gimpel, Sharma, and
  Soricut}{Lan et~al\mbox{.}}{[n.d.]}]%
        {Lan2020ALBERT}
\bibfield{author}{\bibinfo{person}{Zhenzhong Lan}, \bibinfo{person}{Mingda
  Chen}, \bibinfo{person}{Sebastian Goodman}, \bibinfo{person}{Kevin Gimpel},
  \bibinfo{person}{Piyush Sharma}, {and} \bibinfo{person}{Radu Soricut}.}
  \bibinfo{year}{[n.d.]}\natexlab{}.
\newblock \showarticletitle{ALBERT: A Lite BERT for Self-supervised Learning of
  Language Representations}. In \bibinfo{booktitle}{\emph{ICLR 2020}}.
\newblock


\bibitem[\protect\citeauthoryear{Larsson, Maire, and Shakhnarovich}{Larsson
  et~al\mbox{.}}{[n.d.]}]%
        {larsson2016learning}
\bibfield{author}{\bibinfo{person}{Gustav Larsson}, \bibinfo{person}{Michael
  Maire}, {and} \bibinfo{person}{Gregory Shakhnarovich}.}
  \bibinfo{year}{[n.d.]}\natexlab{}.
\newblock \showarticletitle{Learning Representations for Automatic
  Colorization}. In \bibinfo{booktitle}{\emph{ECCV 2016}}.
\newblock


\bibitem[\protect\citeauthoryear{Liu, Rogers, Shiau, Kislyuk, Ma, Zhong, Liu,
  and Jing}{Liu et~al\mbox{.}}{[n.d.]}]%
        {Liu2017}
\bibfield{author}{\bibinfo{person}{David~C. Liu}, \bibinfo{person}{Stephanie
  Rogers}, \bibinfo{person}{Raymond Shiau}, \bibinfo{person}{Dmitry Kislyuk},
  \bibinfo{person}{Kevin~C. Ma}, \bibinfo{person}{Zhigang Zhong},
  \bibinfo{person}{Jenny Liu}, {and} \bibinfo{person}{Yushi Jing}.}
  \bibinfo{year}{[n.d.]}\natexlab{}.
\newblock \showarticletitle{Related Pins at Pinterest: The Evolution of a
  Real-World Recommender System}. In \bibinfo{booktitle}{\emph{WWW 2017}}.
\newblock


\bibitem[\protect\citeauthoryear{Ma, Zhou, Yang, Cui, Wang, and Zhu}{Ma
  et~al\mbox{.}}{[n.d.]}]%
        {ma2020kdd}
\bibfield{author}{\bibinfo{person}{Jianxin Ma}, \bibinfo{person}{Chang Zhou},
  \bibinfo{person}{Hongxia Yang}, \bibinfo{person}{Peng Cui},
  \bibinfo{person}{Xin Wang}, {and} \bibinfo{person}{Wenwu Zhu}.}
  \bibinfo{year}{[n.d.]}\natexlab{}.
\newblock \showarticletitle{Disentangled Self-Supervision in Sequential
  Recommenders}. In \bibinfo{booktitle}{\emph{KDD 2020}}.
\newblock


\bibitem[\protect\citeauthoryear{Mark~Levy}{Mark~Levy}{[n.d.]}]%
        {Mark2010}
\bibfield{author}{\bibinfo{person}{Klaas~Bosteels Mark~Levy}.}
  \bibinfo{year}{[n.d.]}\natexlab{}.
\newblock \showarticletitle{Music Recommendation and the Long Tail}. In
  \bibinfo{booktitle}{\emph{1st Workshop On Music Recommendation And Discovery
  (WOMRAD), ACM RecSys, 2010}}.
\newblock


\bibitem[\protect\citeauthoryear{Mehrotra, Lalmas, Kenney, Lim-Meng, and
  Hashemian}{Mehrotra et~al\mbox{.}}{[n.d.]}]%
        {mounia_www_2019}
\bibfield{author}{\bibinfo{person}{Rishabh Mehrotra}, \bibinfo{person}{Mounia
  Lalmas}, \bibinfo{person}{Doug Kenney}, \bibinfo{person}{Thomas Lim-Meng},
  {and} \bibinfo{person}{Golli Hashemian}.} \bibinfo{year}{[n.d.]}\natexlab{}.
\newblock \showarticletitle{Jointly Leveraging Intent and Interaction Signals
  to Predict User Satisfaction with Slate Recommendations}. In
  \bibinfo{booktitle}{\emph{WWW 2019}}.
\newblock


\bibitem[\protect\citeauthoryear{Milojevi\'{c}}{Milojevi\'{c}}{2010}]%
        {Milojevic2010}
\bibfield{author}{\bibinfo{person}{Sta\v{s}a Milojevi\'{c}}.}
  \bibinfo{year}{2010}\natexlab{}.
\newblock \showarticletitle{Power Law Distributions in Information Science:
  Making the Case for Logarithmic Binning}.
\newblock \bibinfo{journal}{\emph{J. Am. Soc. Inf. Sci. Technol.}}
  \bibinfo{volume}{61}, \bibinfo{number}{12} (\bibinfo{date}{Dec.}
  \bibinfo{year}{2010}), \bibinfo{pages}{2417–2425}.
\newblock
\showISSN{1532-2882}


\bibitem[\protect\citeauthoryear{Naumov, Mudigere, Shi, Huang, Sundaraman,
  Park, Wang, Gupta, Wu, Azzolini, Dzhulgakov, Mallevich, Cherniavskii, Lu,
  Krishnamoorthi, Yu, Kondratenko, Pereira, Chen, Chen, Rao, Jia, Xiong, and
  Smelyanskiy}{Naumov et~al\mbox{.}}{2019}]%
        {maxim2019}
\bibfield{author}{\bibinfo{person}{Maxim Naumov}, \bibinfo{person}{Dheevatsa
  Mudigere}, \bibinfo{person}{Hao{-}Jun~Michael Shi}, \bibinfo{person}{Jianyu
  Huang}, \bibinfo{person}{Narayanan Sundaraman}, \bibinfo{person}{Jongsoo
  Park}, \bibinfo{person}{Xiaodong Wang}, \bibinfo{person}{Udit Gupta},
  \bibinfo{person}{Carole{-}Jean Wu}, \bibinfo{person}{Alisson~G. Azzolini},
  \bibinfo{person}{Dmytro Dzhulgakov}, \bibinfo{person}{Andrey Mallevich},
  \bibinfo{person}{Ilia Cherniavskii}, \bibinfo{person}{Yinghai Lu},
  \bibinfo{person}{Raghuraman Krishnamoorthi}, \bibinfo{person}{Ansha Yu},
  \bibinfo{person}{Volodymyr Kondratenko}, \bibinfo{person}{Stephanie Pereira},
  \bibinfo{person}{Xianjie Chen}, \bibinfo{person}{Wenlin Chen},
  \bibinfo{person}{Vijay Rao}, \bibinfo{person}{Bill Jia},
  \bibinfo{person}{Liang Xiong}, {and} \bibinfo{person}{Misha Smelyanskiy}.}
  \bibinfo{year}{2019}\natexlab{}.
\newblock \showarticletitle{Deep Learning Recommendation Model for
  Personalization and Recommendation Systems}.
\newblock \bibinfo{journal}{\emph{CoRR}}  \bibinfo{volume}{abs/1906.00091}
  (\bibinfo{year}{2019}).
\newblock


\bibitem[\protect\citeauthoryear{Niu, Caverlee, and Lu}{Niu
  et~al\mbox{.}}{[n.d.]}]%
        {npr_wsdm_2018}
\bibfield{author}{\bibinfo{person}{Wei Niu}, \bibinfo{person}{James Caverlee},
  {and} \bibinfo{person}{Haokai Lu}.} \bibinfo{year}{[n.d.]}\natexlab{}.
\newblock \showarticletitle{Neural Personalized Ranking for Image
  Recommendation}. In \bibinfo{booktitle}{\emph{WSDM 2018}}.
\newblock


\bibitem[\protect\citeauthoryear{Noroozi and Favaro}{Noroozi and
  Favaro}{[n.d.]}]%
        {noroozi2016unsupervised}
\bibfield{author}{\bibinfo{person}{Mehdi Noroozi} {and} \bibinfo{person}{Paolo
  Favaro}.} \bibinfo{year}{[n.d.]}\natexlab{}.
\newblock \showarticletitle{Unsupervised Learning of Visual Representations by
  Solving Jigsaw Puzzles}. In \bibinfo{booktitle}{\emph{ECCV 2016}}.
\newblock


\bibitem[\protect\citeauthoryear{Okura, Tagami, Ono, and Tajima}{Okura
  et~al\mbox{.}}{[n.d.]}]%
        {Okura2017EmbeddingbasedNR}
\bibfield{author}{\bibinfo{person}{Shumpei Okura}, \bibinfo{person}{Yukihiro
  Tagami}, \bibinfo{person}{Shingo Ono}, {and} \bibinfo{person}{Akira Tajima}.}
  \bibinfo{year}{[n.d.]}\natexlab{}.
\newblock \showarticletitle{Embedding-Based News Recommendation for Millions of
  Users}. In \bibinfo{booktitle}{\emph{KDD 2017}}.
\newblock


\bibitem[\protect\citeauthoryear{Volkovs, Yu, and Poutanen}{Volkovs
  et~al\mbox{.}}{[n.d.]}]%
        {Volkovs2017}
\bibfield{author}{\bibinfo{person}{Maksims Volkovs}, \bibinfo{person}{Guangwei
  Yu}, {and} \bibinfo{person}{Tomi Poutanen}.}
  \bibinfo{year}{[n.d.]}\natexlab{}.
\newblock \showarticletitle{DropoutNet: Addressing Cold Start in Recommender
  Systems}.
\newblock In \bibinfo{booktitle}{\emph{Neurips 2017}}.
\newblock


\bibitem[\protect\citeauthoryear{Wu, Xiong, Yu, and Lin}{Wu
  et~al\mbox{.}}{2018}]%
        {wu2018unsupervised}
\bibfield{author}{\bibinfo{person}{Zhirong Wu}, \bibinfo{person}{Yuanjun
  Xiong}, \bibinfo{person}{Stella Yu}, {and} \bibinfo{person}{Dahua Lin}.}
  \bibinfo{year}{2018}\natexlab{}.
\newblock \showarticletitle{Unsupervised Feature Learning via Non-Parametric
  Instance-level Discrimination}.
\newblock \bibinfo{journal}{\emph{CoRR}}  \bibinfo{volume}{abs/1805.01978}
  (\bibinfo{year}{2018}).
\newblock
\urldef\tempurl%
\url{http://arxiv.org/abs/1805.01978}
\showURL{%
\tempurl}


\bibitem[\protect\citeauthoryear{Xin, Karatzoglou, Arapakis, and Jose}{Xin
  et~al\mbox{.}}{[n.d.]}]%
        {Xin2020SelfSupervisedRL}
\bibfield{author}{\bibinfo{person}{Xin Xin}, \bibinfo{person}{Alexandros
  Karatzoglou}, \bibinfo{person}{I. Arapakis}, {and} \bibinfo{person}{J.
  Jose}.} \bibinfo{year}{[n.d.]}\natexlab{}.
\newblock \showarticletitle{Self-Supervised Reinforcement Learning for
  Recommender Systems}.
\newblock \bibinfo{journal}{\emph{SIGIR 2020}} (\bibinfo{year}{[n.\,d.]}).
\newblock


\bibitem[\protect\citeauthoryear{Yang, Yuan, Cer, Kong, Constant, Pilar, Ge,
  Sung, Strope, and Kurzweil}{Yang et~al\mbox{.}}{2018}]%
        {yang-etal-2018-learning}
\bibfield{author}{\bibinfo{person}{Yinfei Yang}, \bibinfo{person}{Steve Yuan},
  \bibinfo{person}{Daniel Cer}, \bibinfo{person}{Sheng-yi Kong},
  \bibinfo{person}{Noah Constant}, \bibinfo{person}{Petr Pilar},
  \bibinfo{person}{Heming Ge}, \bibinfo{person}{Yun-Hsuan Sung},
  \bibinfo{person}{Brian Strope}, {and} \bibinfo{person}{Ray Kurzweil}.}
  \bibinfo{year}{2018}\natexlab{}.
\newblock \showarticletitle{Learning Semantic Textual Similarity from
  Conversations}. In \bibinfo{booktitle}{\emph{Proceedings of The Third
  Workshop on Representation Learning for {NLP}}}. \bibinfo{publisher}{ACL},
  \bibinfo{pages}{164--174}.
\newblock


\bibitem[\protect\citeauthoryear{Yi, Yang, Hong, Cheng, Heldt, Kumthekar, Zhao,
  Wei, and Chi}{Yi et~al\mbox{.}}{[n.d.]}]%
        {yi2019samplingbias}
\bibfield{author}{\bibinfo{person}{Xinyang Yi}, \bibinfo{person}{Ji Yang},
  \bibinfo{person}{Lichan Hong}, \bibinfo{person}{Derek~Zhiyuan Cheng},
  \bibinfo{person}{Lukasz Heldt}, \bibinfo{person}{Aditee Kumthekar},
  \bibinfo{person}{Zhe Zhao}, \bibinfo{person}{Li Wei}, {and}
  \bibinfo{person}{Ed Chi}.} \bibinfo{year}{[n.d.]}\natexlab{}.
\newblock \showarticletitle{Sampling-Bias-Corrected Neural Modeling for Large
  Corpus Item Recommendations}. In \bibinfo{booktitle}{\emph{RecSys 2019}}.
\newblock


\bibitem[\protect\citeauthoryear{Zhai, Kislyuk, Jing, Feng, Tzeng, Donahue, Du,
  and Darrell}{Zhai et~al\mbox{.}}{[n.d.]}]%
        {Zhai2017}
\bibfield{author}{\bibinfo{person}{Andrew Zhai}, \bibinfo{person}{Dmitry
  Kislyuk}, \bibinfo{person}{Yushi Jing}, \bibinfo{person}{Michael Feng},
  \bibinfo{person}{Eric Tzeng}, \bibinfo{person}{Jeff Donahue},
  \bibinfo{person}{Yue~Li Du}, {and} \bibinfo{person}{Trevor Darrell}.}
  \bibinfo{year}{[n.d.]}\natexlab{}.
\newblock \showarticletitle{Visual Discovery at Pinterest}. In
  \bibinfo{booktitle}{\emph{WWW 2017}}.
\newblock


\bibitem[\protect\citeauthoryear{Zhang, Yu, Kumar, and Chang}{Zhang
  et~al\mbox{.}}{[n.d.]}]%
        {xu2017spreadout}
\bibfield{author}{\bibinfo{person}{Xu Zhang}, \bibinfo{person}{Felix~X. Yu},
  \bibinfo{person}{Sanjiv Kumar}, {and} \bibinfo{person}{Shih{-}Fu Chang}.}
  \bibinfo{year}{[n.d.]}\natexlab{}.
\newblock \showarticletitle{Learning Spread-Out Local Feature Descriptors}. In
  \bibinfo{booktitle}{\emph{ICCV 2017}}.
\newblock


\bibitem[\protect\citeauthoryear{Zhao, Hong, Wei, Chen, Nath, Andrews,
  Kumthekar, Sathiamoorthy, Yi, and Chi}{Zhao et~al\mbox{.}}{[n.d.]}]%
        {zhe19watchnext}
\bibfield{author}{\bibinfo{person}{Zhe Zhao}, \bibinfo{person}{Lichan Hong},
  \bibinfo{person}{Li Wei}, \bibinfo{person}{Jilin Chen},
  \bibinfo{person}{Aniruddh Nath}, \bibinfo{person}{Shawn Andrews},
  \bibinfo{person}{Aditee Kumthekar}, \bibinfo{person}{Maheswaran
  Sathiamoorthy}, \bibinfo{person}{Xinyang Yi}, {and} \bibinfo{person}{Ed
  Chi}.} \bibinfo{year}{[n.d.]}\natexlab{}.
\newblock \showarticletitle{Recommending What Video to Watch next: A Multitask
  Ranking System}. In \bibinfo{booktitle}{\emph{RecSys 2019}}.
\newblock


\bibitem[\protect\citeauthoryear{Zhou, Wang, Zhao, Zhu, Wang, Zhang, yuan Wang,
  and Wen}{Zhou et~al\mbox{.}}{[n.d.]}]%
        {Zhou2020S3RecSL}
\bibfield{author}{\bibinfo{person}{Kun Zhou}, \bibinfo{person}{Haibo Wang},
  \bibinfo{person}{Wayne~Xin Zhao}, \bibinfo{person}{Yutao Zhu},
  \bibinfo{person}{Sirui Wang}, \bibinfo{person}{Fuzheng Zhang},
  \bibinfo{person}{Zhong yuan Wang}, {and} \bibinfo{person}{Jirong Wen}.}
  \bibinfo{year}{[n.d.]}\natexlab{}.
\newblock \showarticletitle{S3-Rec: Self-Supervised Learning for Sequential
  Recommendation with Mutual Information Maximization}.
\newblock \bibinfo{journal}{\emph{CIKM 2020}} (\bibinfo{year}{[n.\,d.]}).
\newblock


\end{thebibliography}

\newpage

\appendix
\section{Appendix}
\addcontentsline{toc}{section}{Appendices}
\renewcommand{\thesubsection}{\Alph{subsection}}

\subsubsection{Dataset Statistics} \label{appendix:data_stats}

\begin{figure}
        \includegraphics[width=.48\columnwidth]{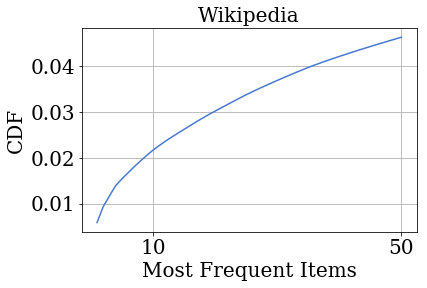}
        \includegraphics[width=.48\columnwidth]{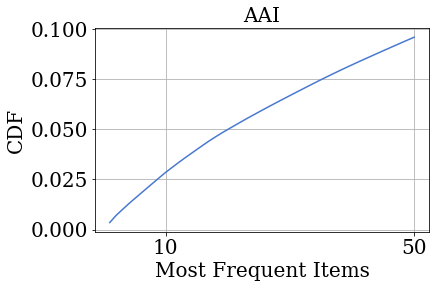}
    \caption{CDF of most frequent items in the Wikipedia and AAI datasets. The distribution is extremely dominated by popular items. For example, the top 50 items out of the 2.4M items already constitute $10\%$ of data in the AAI dataset.}
    \label{fig:dataset}
\end{figure}

Table \ref{tab:dataset} shows some basic stats for the Wikipedia and AAI datasets. Figure ~\ref{fig:dataset} shows the CDF of most frequent items for the two datasets, indicating a highly skewed data distribution. For example, the top 50 items in the AAI dataset collectively appeared roughly 10\% in the training data. If we consider a naive baseline (i.e., \textit{TopPopular} recommender \cite{dacrema2019really}) that recommends the most frequent top-K items for every query, the $CDF$ of the $K$-th frequent item essentially represents the $Recall@K$ metric of such baseline. This suggests a naive TopPopular recommender achieves $Recall@50 \approx 0.1$ for AAI and $Recall@50 \approx 0.05$ for Wikipedia. We present that all the proposed methods outperform this baseline by a large margin in Section \ref{sec:exp}.

%\subsection{Experiment Setup}\label{sec_appendix:exp_setup}
\subsubsection{Head and Tail Item Evaluation Dataset}\label{sec_appendix:head_and_tail_data}
We partition the full test dataset based on popularity of the groundtruth item. For the AAI test dataset, the \textit{Head} dataset consists examples where the groundtruth items are in the top $10\%$ most frequent items, and the rest of the test examples are treated as \textit{tail}. For Wikipedia, we follow the data partitions in ~\cite{krichene2018efficient}, where test examples containing items not included in the training set are treated as \textsl{Tail}, and the rest test examples are treated as \textsl{Head}.

\subsubsection{Visualization of Learned Embeddings}
\label{sec:visualization}

Besides better model performance, we expect the representations learned with SSL to have better quality than the counterparts without SSL. To verify our hypothesis, we take the app embeddings learned in the models trained on AAI dataset, and plot them using t-SNE plot in Figure~\ref{fig:visualization}. Apps from different categories are plotted in different colors, as illustrated in the legends in Figure~\ref{fig:visualization}. Compared to the apps in (Figure~\ref{fig:viz_baseline}), the apps in the best SSL model (Figure~\ref{fig:viz_cfd}) tend to group much better with similar apps in the same category, and the separation of different category looks much more clear. For example, we could see that the ``Sports \& Recreation'' apps (in red) are mixed together with ``Law \& Government'' and ``Travel'' apps in Figure~\ref{fig:viz_baseline}. While in Figure~\ref{fig:viz_cfd}, we clearly see the 4 categories of apps grouped together among themselves. This indicates that the representations learned with SSL carry more semantic information, and is also why SSL leads to better model performance in our experiments.

\begin{table}[]
    \centering
    \begin{tabular}{c|c|c|c}
         Dataset & {\#} queries & {\#} items & {\#} examples \\\hline
         Wikipedia &  5.3M & 5.3M & 490M \\\hline
         AAI & 2.4M & 2.4M & 1B
    \end{tabular}
    \caption{Corpus sizes of the Wikipedia and the AAI datasets.}
    \label{tab:dataset}
\end{table}

\end{document}